\PassOptionsToPackage{colorlinks=true,linkcolor=blue,citecolor=red}{hyperref}
\PassOptionsToPackage{dvipsnames}{xcolor}
\documentclass[sigconf,authorversion,nonacm]{acmart}
\usepackage{microtype}
\usepackage{graphicx}
\usepackage{subfigure}
\usepackage{booktabs} 
\usepackage{amsmath}
\usepackage{enumitem}
\usepackage[ruled,algo2e]{algorithm2e}

\SetCommentSty{mycommfont}
\usepackage{setspace}
\usepackage{wrapfig}
\usepackage{bm}

\usepackage{colortbl}

\usepackage{hyperref}

\newcommand{\fedrule}{\textbf{\texttt{FedRule}}\@\xspace}
\newcommand{\graphrule}{\textbf{\texttt{GraphRule}}\@\xspace}
\definecolor{LightCyan}{rgb}{0.67,0.9,1}
\newcommand{\mycc}{\cellcolor{LightCyan}}

\usepackage{adjustbox}
\usepackage{textcomp}

\begin{document}
\title[FedRule: Federated Rule Recommendation System with Graph Neural Networks]{FedRule: Federated Rule Recommendation System \\ with Graph Neural Networks}

\author{Yuhang Yao}
\authornote{Both authors contributed equally to this research.}
\authornote{This work was done during Yuhang Yao's internship at Wyze Labs.}
\affiliation{%
  \institution{Electrical and Computer Engineering}
  \institution{Carnegie Mellon University}
  \country{}
}
\email{yuhangya@andrew.cmu.edu}

\author{Mohammad Mahdi Kamani}
\authornotemark[1]
\affiliation{%
  \institution{AI Team}
  \institution{Wyze Labs}
  \country{}
}
\email{mkamani@wyze.com}

\author{Zhongwei Cheng }
\affiliation{%
  \institution{AI Team}
  \institution{Wyze Labs}
  \country{}
}
\email{zcheng@wyze.com}

\author{Lin Chen}
\affiliation{%
  \institution{AI Team}
  \institution{Wyze Labs}
  \country{}
}
\email{lchen@wyze.com}

\author{Carlee Joe-Wong}
\affiliation{%
  \institution{Electrical and Computer Engineering}
  \institution{Carnegie Mellon University}
  \country{}
}
\email{cjoewong@andrew.cmu.edu}

\author{Tianqiang Liu}
\affiliation{%
  \institution{AI Team}
  \institution{Wyze Labs}
  \country{}
}
\email{tliu@wyze.com}



\begin{abstract}
   Much of the value that IoT (Internet-of-Things) devices bring to ``smart'' homes lies in their ability to automatically trigger other devices' actions: for example, a smart camera triggering a smart lock to unlock a door. Manually setting up these rules for smart devices or applications, however, is time-consuming and inefficient. Rule recommendation systems can automatically suggest rules for users by learning which rules are popular based on those previously deployed (e.g., in others' smart homes). Conventional recommendation formulations require a central server to record the rules used in many users' homes, which compromises their privacy and leaves them vulnerable to attacks on the central server's database of rules.  Moreover, these solutions typically leverage generic user-item matrix methods that do not fully exploit the structure of the rule recommendation problem. In this paper, we propose a new rule recommendation system, dubbed as \fedrule, to address these challenges. One graph is constructed per user upon the rules s/he is using, and the rule recommendation is formulated as a link prediction task in these graphs. This formulation enables us to design a federated training algorithm that is able to keep users' data private. Extensive experiments corroborate our claims by demonstrating that \fedrule has comparable performance as the centralized setting and outperforms conventional solutions.    
\end{abstract}




\settopmatter{printacmref=false}
\maketitle

\section{Introduction}\label{sec:intro}
With the rapid expansion of smart devices and applications in recent years, it becomes imperative to automate the actions of different devices and applications by connecting them together. For example, an occupancy sensor change can trigger a smart thermostat to turn on, or a code merge can trigger software updates. These connections are broadly construed as \textit{rules} between entities in different systems. Setting up these rules between many entities can involve a tedious and challenging search process, especially for new users.  Hence, it is helpful to provide new users with meaningful recommendations by learning from other users' sets of rules. 

\begin{figure*}[t!]
     \centering
     \subfigure[User-Item Matrix Design]{
     \centering
     \includegraphics[width=0.3\textwidth]{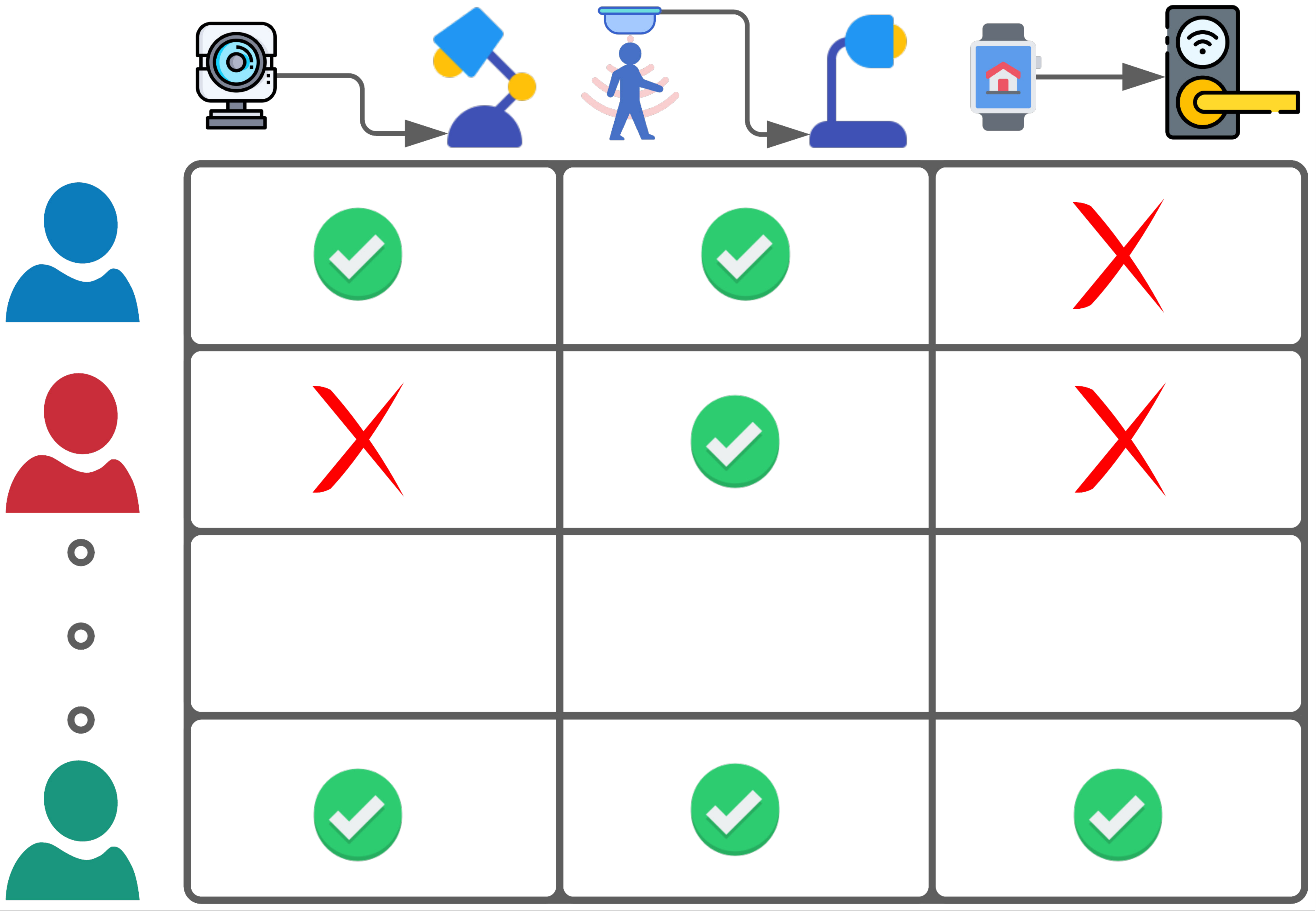}
     \label{fig:user_item_matrix}
     }\hspace{1cm}
     \subfigure[Graph Structure Design]{
     \centering
     \includegraphics[width=0.5\textwidth]{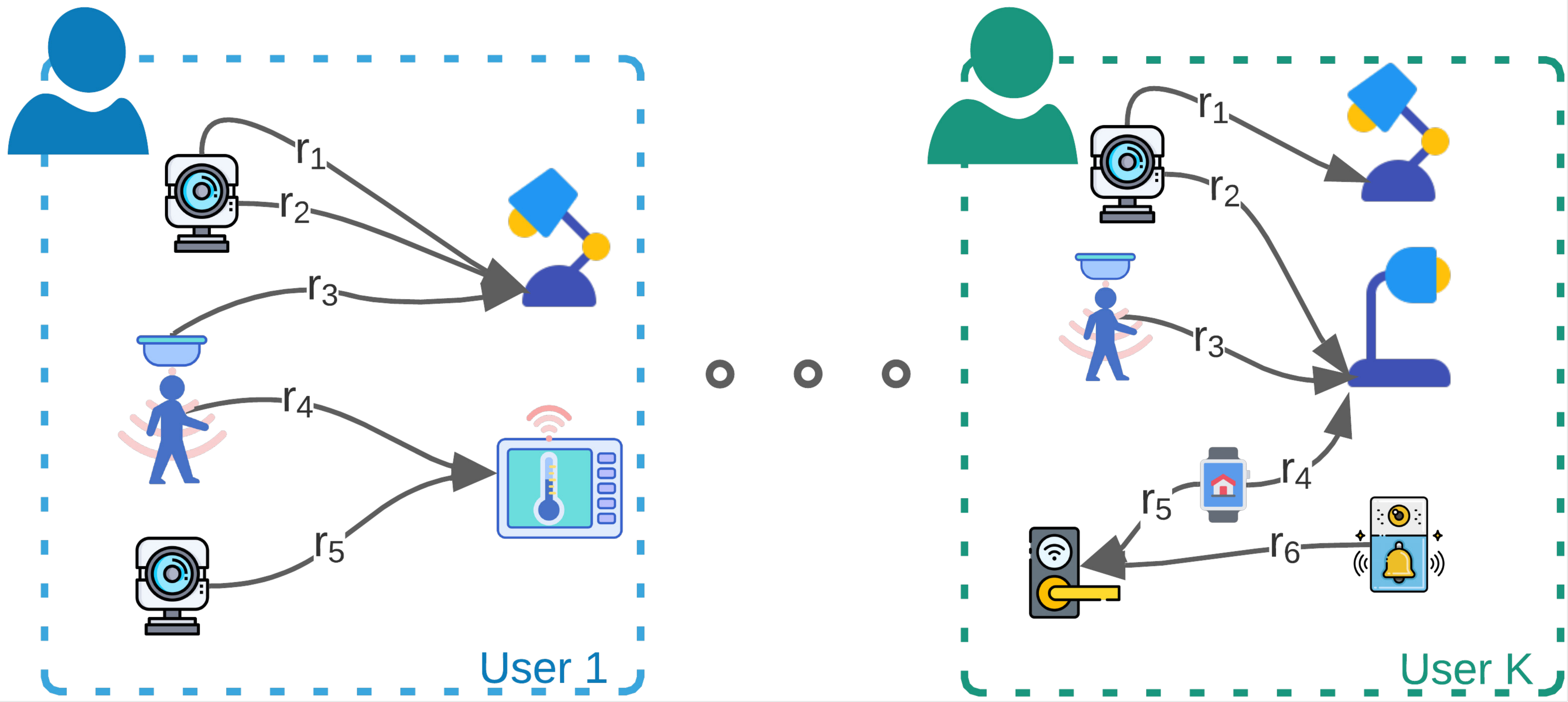}
     \label{fig:rule_device_graph}
     }
     \caption{Illustrations of the conventional structure of recommendation systems based on user-item structure (left), and the proposed graph-based structure for rule recommendation system in a smart home device connections setting (right). In the conventional setting, each rule is considered an item while in the graph structure each rule is represented as an edge between pairs of entity nodes.}\label{fig:intro}
 \end{figure*}
Recommendation systems are growing in various applications from e-commerce~\cite{linden2003amazon} to social networks~\cite{walter2008model} and entertainment industries~\cite{bennett2007netflix,gomez2015netflix}. In most cases, the recommendation problem is formulated as a matrix completion task of a user-item matrix~\cite{corno2019recrules, wu2020learning,berg2017graph} to recommend new items to the users based on previous items chosen by the user. In this way, the recommendation problem is reduced to a dual link between users and items, as depicted in Figure~\ref{fig:user_item_matrix}. Following this setting, we consider each rule as an item for recommending linked rules to users. However, this is not desirable as the structure of entities (\emph{e.g.} the structure of devices or applications owned by a user) on the user side is not considered. Hence, we are not able to distinguish between different instances of an entity type (\emph{e.g.} multiple cameras or multiple motion sensors) to provide meaningful recommendations for each separate entity based on their existing structure.
For instance, in a smart home network, there could be multiple contact sensors connected to a different set of devices in their vicinity. Without considering the existing structure of the connected devices, the recommendation system tends to recommend the same rules for all contact sensors as they are recognized as one general entity instead of several independent ones. Moreover, \emph{privacy} and \emph{security} concerns for training models on users' data in recommendation systems formulated as user-item matrix completion become a great challenge in the IoT system. The user-item formulations require a central server to know all users' rules: these may be sensitive, especially if they involve smart devices' behavior in users' homes.

In this paper, we, first, propose a new rule recommendation framework dubbed as \graphrule , based on the graph structure of entities for each user. In our proposed setup, we create a directed graph for each user based on their available entities (\emph{e.g.} light bulb and contact sensor) as nodes and their rule connections as edges in this graph. Therefore, instead of representing a user's rules as a row in the user-item matrix (each item is an \textcolor{NavyBlue}{entity}-\textcolor{ForestGreen}{rule}-\textcolor{Plum}{entity} triplet, \emph{e.g.} when \textcolor{NavyBlue}{contact sensor} \textcolor{ForestGreen}{is open, turn on} \textcolor{Plum}{light bulb}), we represent each users' rules as a graph that encapsulates the structure of entities and how they are connected together through the available rules.
As an example, in Figure~\ref{fig:rule_device_graph}, for user 1, we have two separate cameras denoted by different nodes in the graph, each connecting to a distinct set of devices by their specific rules. However, it is infeasible to distinguish between these cameras in the user-item structure without separating the rule and entities representations from the entity-rule-entity items in user-item matrices. In fact, in user-item design, we condense entities with the same type into one general entity (\emph{e.g.} camera 1 and camera 2 into the camera) to only emphasize their connections to other entities as rules. Based on our graph structure design, the goal of the rule recommendation system is then to predict the newly formed edges (rules) in the graph, which can be formulated as a graph link prediction problem~\cite{zhang2018link}. The system first learns node embeddings for each entity in the user's graph using graph neural network architectures~\cite{scarselli2008graph,velivckovic2017graph}. Based on the learned embeddings, a parametric prediction model can be used to estimate the probability of different rules connecting each pair of entities in the graph.

For privacy concerns, unlike the conventional user-item structure, the proposed graph structure can be easily distributed across different users for training the main recommendation model locally. Indeed, by casting the rule recommendation problem as a link prediction task across graphs for different users, we can employ privacy-preserving training setups such as federated learning~\cite{mcmahan2017communication}, which allows data to remain at the user's side. Moreover, federated training methods are generally iterative, and can thus easily adapt the model to new users' data instead of having to retrain the whole model from scratch, as is common in previous approaches for recommendation systems.

There has been a surge in applications of federated learning in the training of machine learning models recently~\cite{mcmahan2017communication,kairouz2019advances,bonawitz2019towards,haddadpour2019local,deng2020adaptive,deng2020distributionally}, with some applications in recommendation systems~\cite{wu2021fedgnn,chen2020fede}. However, most of the proposals for recommendation tasks in federated learning are for \emph{cross-silo} setups (\emph{i.e.}, data could be distributed across pre-defined silos) based on user-item matrix completion, and not for \emph{cross-device} settings~\cite{kairouz2019advances} (\emph{i.e.}, with limited data at each user, where here "device" represents one user and its entities), as in our rule recommendation task.

Our cross-device setting with the proposed graph structure, however, introduces new challenges: the amount of data on each individual user is very limited (as each user generally has only a few entities, e.g., a few smart home devices) and the local data is not independent and identically distributed (non-IID). The non-IID data distribution with a small sample size severely increases the variance of gradients among clients. Hence, when simply applying federated training methods, like FedAvg~\cite{mcmahan2017communication, he2021fedgraphnn,haddadpour2019convergence}, the model will not converge easily and cannot have comparable performance on par with the centralized training. To overcome the issue of severely non-IID data distribution across clients, we use two control parameters in local machines for each client to correct the gradients for different parts of our model and avoid drifting too much from the average model. Given the limited data of users, we sample negative edges (rules that do not exist) to balance the numbers of positive and negative edges. With these solutions, we then introduce \fedrule, a federated rule recommendation system with graph neural networks. \fedrule can learn the representation and link prediction models of the recommendation system at the user level with a decent convergence rate using variance reduction techniques with control parameters while preserving the privacy of users' data.  With negative sampling and variance reduction to alleviate the effects of limited and non-IID data, this system enables us to learn a better model in a graph-based cross-device setting over user-item-based settings. In our empirical results, we show the effectiveness of our \fedrule over other approaches.

Our main contributions are:
\begin{itemize}[nosep]
    \item We propose a new rule recommendation framework, called \graphrule, based on the graph structure of current users by representing rules as edges between entity nodes over conventional user-item structure. 
    \item We propose an effective federated learning system, called \fedrule, which effectively learns from limited non-IID data on each user's rules with fast convergence and preserving the privacy of users' data.
    \item Extensive empirical investigation shows the effectiveness of both the proposed structure for the \graphrule and the \fedrule algorithms. Our \fedrule algorithm is able to achieve the same performance as the centralized version (\graphrule) due to its variance reduction mechanism.
\end{itemize}
\section{Related Work}\label{sec:related} 
\paragraph{Rule Recommendation Systems} 
Traditional recommendation systems are mainly based on matrix factorization techniques~\cite{koren2009matrix, bokde2015matrix} and aim to match users to a set of items they are interested in, based on the features of users and items as well as prior data expressing users' interests in a given set of items. Recently, graph neural networks have been widely adopted by many recommendation systems to model this relationship as a bipartite graph structure. GCMC~\cite{berg2017graph} provides a graph-based auto-encoder framework for matrix completion problems. Given different relations between users and items, R-GCNs~\cite{schlichtkrull2018modeling} models relational data with graph convolutional networks and HetGNN~\cite{zhang2019heterogeneous} is to deal with heterogeneous graphs with features of users and items. RotatE~\cite{sun2019rotate} uses knowledge graph embedding by relational rotation in complex number space.

With the development of the smart home technology~\cite{yu2020learning}, rule recommendation systems for IoT devices exhibit great practical and research potential. IFTTT, a mobile app providing rule execution services for mobile applications and IoT devices, provides open access data with users and rules for analysis ~\cite{ur2016trigger,mi2017empirical}. For recommending rules in IFTTT, RecRule~\cite{corno2019recrules} provides rule recommendation based on a user-rule matrix with a semantic reasoning process to enrich rules with semantic information of rules, and TaKG~\cite{wu2020learning} builds a trigger-action knowledge graph with the user-rule collaborative filtering to recommend rules. 

\paragraph{Graph Neural Networks}
Graph neural networks aim to learn representations of graphically structured data that capture features associated with graph nodes as well as edges between the nodes~\cite{bronstein2017geometric}. Recently graph neural networks, Graph Convolutional Networks (GCN)~\cite{kipf2016semi}, GraphSage~\cite{hamilton2017inductive}, and Graph Attention Networks (GAT)~\cite{velivckovic2017graph}, are proposed and show great performance on various graph learning tasks. A popular task is that of link prediction~\cite{zhang2018link}, i.e., predicting whether a link exists between two nodes in the graph, given information on the presence or absence of links between other pairs of nodes. We show that the rule recommendation problem can be formulated as a link prediction one, and go beyond prior work on GNNs for link prediction by considering a federated solution algorithm that can be learned in a distributed way across multiple users.

\paragraph{Federated Learning}
Federated learning is first proposed in~\cite{mcmahan2017communication}, which allows clients to train a common machine learning model on their collective data while keeping personal data on clients. Instead of uploading data to the server for centralized training, clients iteratively process their local data and share model updates with the server. Weights from a large population of clients are periodically aggregated by the server and combined to create an improved global model. The widely adopted FedAvg algorithm~\cite{mcmahan2017communication} is used on the server to combine client updates and produce a new global model. However, FedAvg may fail to converge if data from different clients follow different distributions (i.e., is non-IID), as is often the case in federated learning scenarios, including our rule recommendation system. To deal with the non-IID local data distribution~\cite{zhao2018federated, li2019convergence}, various solutions such as SCAFFOLD~\cite{karimireddy2020scaffold}, FedProx~\cite{li2020federated}, and more recently, FedGate~\cite{haddadpour2021federated} are proposed to reduce the variance of data distribution and speed up the learning. 

Recent work proposes federated learning algorithms specialized to graph neural networks, and the model aims to predict characteristics of the graph, such as link prediction, node classification, and graph classification. FedGraphNN~\cite{he2021fedgraphnn} provides a federated learning system and benchmark for graph neural networks, which mainly focuses on the system deployment and only uses FedAvg for the training process. Our work, however, focuses on learning techniques to handle the challenges of limited and non-IID data in rule recommendation problems.

There are also algorithms that aim to use federated learning to make recommendations. FedMF~\cite{chai2020secure} provides secure federated matrix factorization. FCF~\cite{ammad2019federated} provides federated collaborative filtering for privacy-preserving personalized recommendation systems. \citet{wu2021fedgnn} uses a federated heterogeneous graph neural network to model user-item relationships. All these federated recommendation systems use the user-item matrix formulation, which requires communication among clients to exchange user information and is not fully federated, potentially raising privacy and security concerns. By re-formulating the rule recommendation problem to a link prediction task in a graph, we eliminate the need for communication among clients, helping to preserve user privacy and reducing the overhead of communication during training.

\section{Federated Rule Recommendation}\label{sec:fedrule}
Recommendation systems often rely on gathering data from all users and their interactions with items or other users to learn prediction models. This procedure poses a great threat to user privacy, as seen in the Netflix Recommendation challenge, where researchers were able to identify individuals by matching Netflix profiles with those of IMDB profiles~\cite{narayanan2006break}. Hence, it is of paramount importance to consider privacy-preserving systems for recommendation services. Federated Learning~\cite{mcmahan2017communication} provides a viable solution for training such models without the need to move users' data to a server. Thus, it can be employed in training these models on user-item or user-user interaction data.

The main challenge that hinders the applications of federated learning in recommendation systems is the structure of data in such systems, which cannot be distributed easily. For instance, matrix factorization or even recent algorithms such as Graph Convolutional Matrix Completion (GCMC)~\cite{berg2017graph} require access to the full user-item matrix they intend to complete. Although some distributed variations of these algorithms have been proposed~\cite{gemulla2011large,liu2010distributed}, they are mostly focused on computational efficiency in a structured homogeneous distribution of data. Thus, it addresses neither user privacy concerns nor heterogeneity of data across the users' network. For the rule recommendation problem, we address these concerns by proposing a federated learning setup based on the graph structure of rules. Despite our focus on rule recommendation with graph link prediction, the proposed learning approach can evidently be applied to other tasks such as graph classification~\cite{zhang2019hierarchical} and node classification~\cite{kipf2016semi}.

\subsection{Graph Rule Recommendation} \label{sec:rule_recommendation}
In common recommendation systems, the goal is to predict unknown links between users and items. However, in rule recommendation problems we want to suggest how users can connect two or more entities (\emph{e.g.} smart devices) together. In previous approaches, we are limited to considering multiple entities' connections as the connections of one abstract entity to create a user-item matrix (with items as rules). In this case, the item is a rule between those entities, and hence, we are losing information on the graph structure of entities and their connections.

In the rule recommendation problem, similar to the graph link prediction problem~\cite{zhang2018link}, we need to consider the relationship between different entities as well. Hence, instead of representing the connections as a row in a 2D matrix of user-item relationships, we have a graph of entities' connections for each user (see Figure~\ref{fig:intro}). The graph also enables us to represent entities with the same type (\emph{i.e.}, Cameras) as separate entity nodes in the graph. In addition, in this setting, compared to a simple graph, instead of binary connections between different entities, we can have multiple types of edge connections. This will enable us to represent different types of connections as different rules between entities. Using this structure, we can leverage advancements in the graph neural network domain to learn a better model for recommending new rules to the users.

Therefore, given the set of users $\mathcal{U}$ and the set of connection types $\mathcal{R}$, for each user $u_k\in\mathcal{U}$ in our setting, we create a graph of entities denoted by $\mathcal{G}_k=\left(\mathcal{V}_k,\mathcal{X}_k, \mathcal{E}_k\right)$, where $\mathcal{V}_k$ is a set of nodes (entities), $\mathcal{X}_k$ is the set of nodes' (entities') features (\emph{e.g.}, types of entities), and $\mathcal{E}_k$ is the set of connections between nodes in this graph. $\mathcal{X}_k$ can be represented as $\{\mathbf{x}_v, \forall v \in \mathcal{V}_k\}$, where $\mathbf{x}_v$ represents the feature vector of node $v$. Each connection in $\mathcal{E}_k$ can be represented as $(v_i, v_j, r)$, which means there is an edge (rule) between node $v_i$ and node $v_j$ with connection type $r\in \mathcal{R}$ (e.g. "is open, turn on" in "when the contact sensor is open turn on the light"). The goal of the recommendation system in this problem is to learn an embedding model $\bm{\theta}$ for nodes in the graph of user $u_k$ and a predictor $\bm{\phi}$ that can use the node embeddings to estimate the following probability:
\begin{equation}
    \mathbb{P}\big((v_i, v_j, r)\in \mathcal{E}_k | \bm{\theta}, \bm{\phi}; \mathcal{G}_k \big). \label{eq:prob}
\end{equation}
Based on this probability we can then recommend new edges to the users.

\subsection{Model}
For each client, our rule recommendation model has two parts: a Graph Neural Network (GNN) to calculate the node embeddings and a predictor to predict the connections between nodes (the types of rules) as well as their probabilities. As discussed above, each client $u_k$ has a graph $\mathcal{G}_k$ to represent the ground truth connections between different nodes.

\subsubsection{GNN}
We choose a two-layer graph neural network to get the embedding of nodes. Algorithm~\ref{Alg:graphsage} shows the GraphSage~\cite{hamilton2017inductive} model that we used. In each GraphSage layer $l$, at first the differentiable aggregator function $\mathsf{AGG}(\{\mathbf{h}^{l}_{v_n}, \forall v_n \in \mathcal{N}(v)\})$ aggregates information from node neighbors, where $\mathbf{h}^{l}_{v_n}$ denotes the hidden state of node $v_n$ and $\mathcal{N}(v)$ denotes the set of neighbors of node $v$. We use the mean operator as the aggregator function which takes the element-wise mean of the hidden states. Then, weight matrices $\bm{\theta}^1$ and $\bm{\theta}^2$ are used to propagate information between different layers of the model. After the two-layer GNN, we can then get the node embeddings $\mathbf{z}_v^k$ for all $v \in \mathcal{V}_k$.

\begin{figure*}[t!]
    \centering
    \includegraphics[width = 0.9\textwidth]{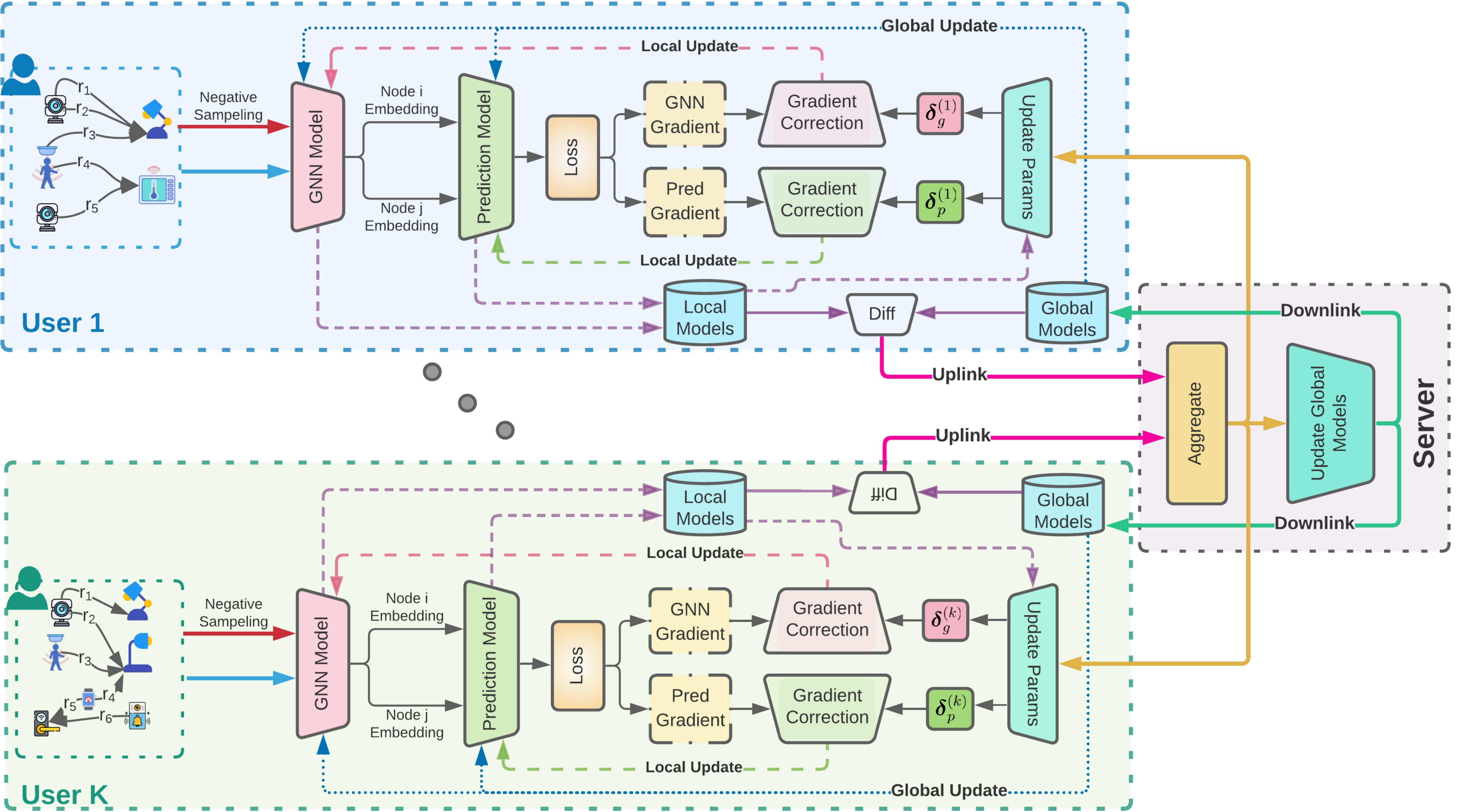}
    \caption{Overview of \fedrule System Architecture Design.}
    \label{fig:fedrulesys}
\end{figure*}

\begin{algorithm2e}[t]
\DontPrintSemicolon
\caption{\texttt{GraphSage} Node Embedding Generation Algorithm for Client $u_k$}
\SetNoFillComment
\label{Alg:graphsage}
\SetKwFor{ForPar}{for}{do in parallel}{end forpar}

\textbf{Inputs}: Graph $\mathcal{G}_k=\left(\mathcal{V}_k,\mathcal{X}_k, \mathcal{E}_k\right)$, where input features $\mathcal{X}_k = \{\mathbf{x}_v, \forall v \in \mathcal{V}_k\}$; Weight matrices $\bm{\theta}^1$ and $\bm{\theta}^2$; Differentiable aggregator function $\mathsf{AGG}\left(.\right)$; Neighborhood function $\mathcal{N}: v \rightarrow 2^{\mathcal{V}_k}$.\\
\textbf{Output}: Node embeddings $\mathbf{z}_v$ for all $v\in \mathcal{V}_k$.\\
    \tcp{Node Features as Hidden State}
    $\mathbf{h}^0_v \leftarrow \mathbf{x}_v, \forall v \in \mathcal{V}_k$;\\
    \tcp{First GraphSage Layer with ReLU Activation}
    \For{$v\in \mathcal{V}_k$}{
        $\mathbf{h}^1_{\mathcal{N}(v)} \leftarrow \mathsf{AGG}(\{\mathbf{h}^{0}_{v_n}, \forall v_n \in \mathcal{N}(v)\})$;\\
        $\mathbf{h}^1_v \leftarrow \text{ReLU} (\bm{\theta}^1 \cdot \mathsf{CONCAT}(\mathbf{h}^{0}_v, \mathbf{h}^0_{\mathcal{N}(v)}))$;\\
        
    }
    \tcp{Second GraphSage Layer}
    \For{$v\in \mathcal{V}_k$}{
        $\mathbf{h}^2_{\mathcal{N}(v)} \leftarrow \mathsf{AGG}(\{\mathbf{h}^{1}_{v_n}, \forall v_n \in \mathcal{N}(v)\})$;\\
        $\mathbf{h}^2_v \leftarrow \bm{\theta}^2 \cdot \mathsf{CONCAT}(\mathbf{h}^{1}_v, \mathbf{h}^1_{\mathcal{N}(v)})$;\\
        
    }
    $\mathbf{z}_v^k \leftarrow \mathbf{h}^2_v, \forall v \in \mathcal{V}_k$

\end{algorithm2e}

\subsubsection{Predictor}
For each client $u_k$, after we get the node embeddings $\mathcal{Z}_k = \{\mathbf{z}_v^k, \forall v \in \mathcal{V}_k\}$. We then need to predict the edge connection probability $\mathbf{p}_{v_i,v_j}^k \in [0,1]^{|\mathcal{R}|}$ between node $v_i$ and node $v_j$ for all $v_i\in \mathcal{V}_k$ and all $v_j\in \mathcal{V}_k$. In our case, for the predictor model, we use a two-layer fully-connected neural network with ReLU activation for the first layer and the Sigmoid function in the last layer.
The predictor uses $\mathbf{z}_{v_i}^k$ and $\mathbf{z}_{v_j}^k$, embeddings of node $v_i$ and node $v_j$ as input, makes it pass the 2-layer neural network ($\bm{\phi}$) with weight matrices $\bm{\phi}^1$ and $\bm{\phi}^2$, then outputs the link probability of edges.

\subsubsection{Centralized Optimization}
The model can be easily trained in a centralized setting where the centralized server stores all user graphs $\mathcal{G}_k=\left(\mathcal{V}_k, \mathcal{X}_k, \mathcal{E}_k\right)$ for all $u_k \in \mathcal{U}$. We call this centralized training on the graph structure for rule recommendation as \graphrule. User graphs are mainly sparse, meaning most of the connections between nodes are not set yet. Given the number of possible types of edges between nodes, there are only a few positive edges $\mathcal{E}_{k}^{\text{pos}}$ where the positive edge means that there is a formed connection between the two entities. The other possible edges can then be considered as negative edges $\mathcal{E}_{k}^{\text{neg}}$. Considering all negative edges lengthens the training time, so we \emph{sample negative edges} to balance the numbers of positive and negative edges.

We then use binary cross entropy loss including positive and negative edges as our objective function:
\begin{equation}
\begin{aligned}
    \mathcal{L}_k(\mathbb{P}^k,\mathcal{E}_{k}^{\text{pos}}, \mathcal{E}_{k}^{\text{neg}}) &= - \frac{1}{|\mathcal{E}_{k}^{\text{pos}}| + |\mathcal{E}_{k}^{\text{neg}}|} \\
    & \left(\sum_{e \in \mathcal{E}_{k}^{\text{pos}}}\log(p_{e}) +
    \sum_{e \in \mathcal{E}_{k}^{\text{neg}}}\log(1 - p_{e})\right),
\end{aligned}
\end{equation}
where $\mathbb{P}^k \in [0,1]^{|\mathcal{V}_k|\times |\mathcal{V}_k| \times |\mathcal{R}|}$ denotes all pairs of edge connection probability in $\mathcal{G}_k$, and $p_{e} \in [0,1]$ denotes the connection probability of a specific edge $e = (v_i, v_j, r)$.

By denoting the GNN as $\bm{\theta}$ and predictor as $\bm{\phi}$, the loss given user graph $\mathcal{G}_k$ can then be represented as 
\begin{equation}
\mathcal{L}_k (\bm{\theta},\bm{\phi};\mathcal{G}_k) = \mathcal{L}_k(\mathbb{P}^k \left(\bm{\theta},\bm{\phi}\right),\mathcal{E}_{k}^{\text{pos}}, \mathcal{E}_{k}^{\text{neg}}). 
\end{equation}

At each iteration, the centralized server performs gradient descent by calculating the loss
\begin{equation}
\mathcal{L}(\bm{\theta},\bm{\phi};\mathcal{G}_1,..,\mathcal{G}_{|\mathcal{U}|}) = \sum_{u_k \in \mathcal{U}} \mathcal{L}_k (\bm{\theta},\bm{\phi};\mathcal{G}_k),
\end{equation}
then updates the model.

\subsection{FedRule}
Learning the aforementioned models at scale requires gathering the graph data of every user at a server to run the training. The \graphrule training, despite its fast convergence speed, can expose users' private data related to the devices or applications they are using and how they are connected together. Hence, it is important to facilitate a privacy-preserving training procedure to safeguard users' data. Federated learning~\citep{mcmahan2017communication} is the de-facto solution for such purpose in distributed training environments. 

Recently, there have been some proposals to apply federated learning in recommendation problems such as in~\cite{wu2021fedgnn,chen2020fede,peng2021federated}. However, almost all these proposals are for cross-silo federated learning and are not as granular as cross-device federated learning. In this paper, the problem of rule recommendation is formulated so as to be more suited for the cross-device federated learning setup. Although the cross-device federated learning setup is more desirable for the purpose of privacy-preserving algorithms, it makes the training procedure more challenging. The reason behind this is that for the problem of rule recommendation, the size of the data (\emph{i.e.}, the graph structure for each client) is small, and it follows non-IID distribution due to heterogeneous user behaviors. The non-IID data distribution increases the variance of gradients among users and makes the gradient updates coming from different users go in different directions. Hence, the local training in federated learning by averaging the gradients is hard to converge due to misaligned directions of gradients. As it is shown in Section~\ref{sec:exp}, applying FedAvg with GNN on cross-device settings~\citep{mcmahan2017communication, he2021fedgraphnn} like rule recommendation problem can fail in some cases due to the non-IID problem mentioned above. 

We then propose the \fedrule, federated rule recommendation system with graph neural network. The design schema of the system is depicted in Figure~\ref{fig:fedrulesys}. We use negative sampling, which samples negative edges in the graph, to balance the numbers of positive and negative edges. To avoid the non-IID problem, we use two control parameters in local machines for each client to correct the gradients and avoid drifting too much from the average model. \fedrule as presented in Algorithm~\ref{Alg:fedrule} consists of four main parts as follows:
    \paragraph{Local Updates} At the beginning of each local training stage (communication round $c$), clients will get the updated global GNN ($\bm{\theta}_k$) and prediction ($\bm{\phi}_k$) models. Then, in each local iteration $t$, the client's device computes the gradients of models using local data. The gradient for the GNN part is with respect to the graph data ($\mathcal{G}_k$) and for the prediction model is with respect to the set of node embeddings ($\mathcal{Z}_{k}$) from the local graph. With the control parameters described next, the gradients get corrected and then the local models ($\bm{\theta}_k^{(t,c)}, \bm{\phi}_k^{(t,c)}$) get updated using their respective learning rates ($\eta_\theta, \eta_\phi$). Note that since the GNN model is a global representation of nodes and the prediction model is more of a personalized classifier, the learning rates of the models might be different.
    \paragraph{Gradient Correction} For each local iteration, after computing the gradients, we adapt the FedGATE algorithm~\cite{haddadpour2021federated} and use two control parameters ($\bm{\delta}_{\bm{\theta}_k}, \bm{\delta}_{\bm{\phi}_k}$) for GNN model and predictor model respectively. The use of these control parameters will help the training to reduce the variance of convergence, as it can be inferred from the experimental results in Section~\ref{sec:exp} as well. Similar to learning rates, due to the different natures of the models, they might get corrected with different rates using parameters $\lambda_{\bm{\theta}}, \lambda_{\bm{\phi}}$.
    \paragraph{Model Aggregation} After $\tau_k$ local steps in each client's device, we aggregate the models from the devices. To do so, we first compute the difference between the current local models and the starting global models at round $c$, denoted by $\bm{\Delta}_{\bm{\theta}_k}^{(\tau_k,c)}, \bm{\Delta}_{\bm{\phi}_k}^{(\tau_k,c)}$. Then, the server averages over these updates from clients and send back these averages to the clients ($\bm{\Delta}_{\bm{\theta}}^{(c)}, \bm{\Delta}_{\bm{\phi}}^{(c)}$). Also, the server uses these average updates to update the global models that need to be broadcast to the clients in the next round. Secure gradient aggregation methods can be integrated into the system to better protect privacy.
    \paragraph{Parameter Updates} Using the calculated average updates in the previous stage, clients update their local control parameters using the deviation of the local updates from average updates: 
    \begin{align}
        \delta_{\bm{\theta}_k}^{(c+1)}=\delta_{\bm{\theta}_k}^{(c)} + \frac{1}{\eta_{\bm{\theta}}\tau_k}\left(\bm{\Delta}_{\bm{\theta}_k}^{(\tau_k,c)} - \bm{\Delta}_{\bm{\theta}}^{(c)}\right)\nonumber\\
        \delta_{\bm{\phi}_k}^{(c+1)}=\delta_{\bm{\phi}_k}^{(c)} + \frac{1}{\eta_{\bm{\phi}}\tau_k}\left(\bm{\Delta}_{\bm{\phi}_k}^{(\tau_k,c)} - \bm{\Delta}_{\bm{\phi}}^{(c)}\right)\label{eq:control}
    \end{align}

\begin{algorithm2e}[t]
\DontPrintSemicolon
\caption{\texttt{FedRule} Federated Learning for Rule Recommendation Systems}
\SetNoFillComment
\label{Alg:fedrule}
\SetKwFor{ForPar}{for}{do in parallel}{end forpar}

\For{$c=1, \ldots, C$}{
    \ForPar{each client $k\in[K]$}{
         Set $\bm{\theta}_k^{(1,c)}={\boldsymbol{\theta}}^{(c)}$, $\boldsymbol{\phi}_k^{(1,c)}={\boldsymbol{\phi}}^{(c)}$,\\
        \For{ $t=1,\ldots,\tau_k$}{
            Set $\bm{g}_{\bm{\theta}_k}^{(t,c)} = \nabla_{\bm{\theta}_k} \mathcal{L}_k (\bm{\theta}_k^{(t,c)},\bm{\phi}_k^{(t,c)};\mathcal{G}_k)$\\
            Set $\bm{g}_{\bm{\phi}_k}^{(t,c)} = \nabla_{\bm{\phi}_k} \mathcal{L}_k (\bm{\theta}_k^{(t,c)}, \bm{\phi}_k^{(t,c)};\mathcal{Z}_{k})$\\
            \tcp{Correct Gradients}
             $\tilde{\boldsymbol{g}}^{(t,c)}_{\left\{\bm{\theta}_k, \bm{\phi}_k\right\}}={\bm{g}}_{\left\{\bm{\theta}_k, \bm{\phi}_k\right\}}^{(t,c)}-\lambda_{\left\{\bm{\theta}, \bm{\phi}\right\}} \delta_{\left\{\bm{\theta}_k, \bm{\phi}_k\right\}}$\\ 
             \tcp{Update Parameters}
             $\bm{\theta}^{(t+1,c)}_{k}=\bm{\theta}^{(t,c)}_k-\eta_{\theta}~ \tilde{\boldsymbol{g}}^{(t,c)}_{\bm{\theta}_k}$\\
             $\bm{\phi}^{(t+1,c)}_{k}=\bm{\phi}^{(t,c)}_k-\eta_{\phi}~ \tilde{\boldsymbol{g}}^{(t,c)}_{\bm{\phi}_k}$\\
        }
        \tcp{Update Control Parameters}
        send \scalebox{0.8}{$ \bm{\Delta}_{\left\{\bm{\theta}_k,\bm{\phi}_k\right\}}^{(\tau_k,c)} = \left\{\bm{\theta}_k^{(c)},\bm{\phi}_k^{(c)}\right\}-\left\{\bm{\theta}_k^{(\tau_k,c)},\bm{\phi}_k^{(\tau_k,c)}\right\}$} to server and gets $\bm{\Delta}^{(c)}_{\left\{\bm{\theta},\bm{\phi}\right\}}$\\
        Update control parameters using Eq.~(\ref{eq:control})
    }
    \tcp{Server Operations}
    ~~~\tcp{Difference Aggregation}
   $\bm{\Delta}^{(c)}_{\left\{\bm{\theta},\bm{\phi}\right\}}=\frac{1}{K}\sum_{k=1}^K \bm{\Delta}_{\left\{\bm{\theta}_k, \bm{\phi}_k\right\}}^{(\tau_k,c)}$ and {broadcasts} back to clients\\
    ~~~\tcp{Update Global Models} Compute \scalebox{0.85}{$\left\{\bm{\theta}^{(c+1)},\bm{\phi}^{(c+1)}\right\}=\left\{\bm{\theta}^{(c)},\bm{\phi}^{(c)}\right\}-\bm{\Delta}^{(c)}_{\left\{\bm{\theta},\bm{\phi}\right\}}$} and broadcast to local clients \\
}
\end{algorithm2e}

\section{Empirical Evaluation}\label{sec:exp} 
After introducing our recommendation formulation, model, and training algorithm, we empirically evaluate the proposed algorithm. In this section, we first introduce the two datasets used in our experiments and comparison methods, then compare our methods in both the centralized and federated settings with baseline algorithms. After that, we show that our proposal can handle multiple entities of the same type cases efficiently. 
\subsection{Datasets}
\paragraph{Wyze Smart Home Rules Dataset}
To experiment the efficacy of our proposed algorithm, we first use a real-world dataset for smart home devices from the Wyze Labs'\footnote{\url{https://www.wyze.com/}} rule engine. This dataset contains the rules that connect smart devices in different clients' houses. Hence, by nature, the distribution of rules among different clients is non-IID, which is in line with the federated learning setting. We call this dataset the ``\textit{Wyze Smart Home Rules}'' dataset, which contains $76,218$ users with $201,940$ rules. We cleaned the dataset to remove all personally identifiable information (PII) in order to address privacy concerns.
We simplify the current rules into the following form: $<$ trigger entity, trigger-action pairs, action entity$>$, where the trigger-action pairs denote the connection type. For instance, we can connect a smart doorbell to another camera, when by pressing the doorbell we want to power on the camera for recording. Then, the rule format is  $<$ Doorbell, Doorbell Pressed - Power On, Camera$>$. We have $11$ unique entities and $163$ unique trigger-action pairs, resulting in a total of $1,207$ unique rules. The types of entities and trigger-action pairs are shown in Table~\ref{table:wyzedata}.

\paragraph{IFTTT Dataset}
The IFTTT Dataset~\cite{corno2019recrules} is one of the most popular EUD tools. To the best of our knowledge, this is the only publicly available dataset of IF-THEN rules defined and shared by different users. But it does not support entities of the same type. It was obtained by~\citet{ur2016trigger} with a web scrape of the IFTTT platform as of September 2016. The dataset contains $144$ different users with $8,729$ rules. There are $3,020$ types of rules used by users. In the current dataset, we use $53$ unique entities and $132$ unique trigger-action pairs. The types of entities and trigger-action pairs are shown in Table \ref{table:iftttdata}.

\subsection{Experimental Setups} 
\paragraph{Comparing Methods} We compare these methods
\begin{itemize}[nosep]
    \item Matrix Factorization \cite{bokde2015matrix}: Complete user-item matrix by matrix factorization.
    \item GCMC \cite{berg2017graph}: Graph-based auto-encoder framework for matrix completion based on a user-item bipartite graph.
    \item \graphrule: The proposed centralized optimization for graph formulation of rule recommendation.
    \item FedAvg with GNN \cite{mcmahan2017communication, he2021fedgraphnn}: Each user has a user graph and a local model. The server aggregates the local models and uses FedAvg to train the global model. We call this FedGNN.
    \item \fedrule: Our proposed federated rule recommendation algorithm.
\end{itemize}

\paragraph{Experiment Setting}
We use the Adam optimizer with a learning rate of $0.1$ and $100$ training rounds. For federated algorithms, we use $3$ local steps at each communication round, for $300$ total iterations. The dimensions of the hidden states between the two GraphSage layers and the two NN layers are $16$ and the number of possible trigger-action pairs, respectively. For \graphrule, users' graphs are stored in the central server and we do the gradient descent with all users' graphs. For federated methods, we compute the batch gradient descent on each user's graph with $3$ local epochs to train the local models, then the server aggregates the local models to update the global model. We set the hyper-parameter $\lambda =1$ for the FedRule algorithm. For each user, we use 80\% rules set by the user for training and the remaining 20\% rules for testing.

\begin{table}[t]
\centering
\begin{adjustbox}{width=\columnwidth,center}
\centering
\begin{tabular}{|l|}
    \hline
         \textbf{Types of Entities}\\
    \hline
         Camera \\
    
         Chime Sensor \\
    
         Contact Sensor \\
    
         Light \\
    
         Lock \\
    
         Mesh Light \\
    
         Motion Sensor \\
    
         Outdoor Plug \\
    
         Plug \\
    
         Thermostat \\
    
         Outdoor Camera \\
    \hline
    \end{tabular}
\centering
\begin{tabular}{|l|}
    \hline
         \textbf{Types of Trigger-Action Pairs}\\
    \hline
         Open, Power On \\
         
         Open, Power Off\\
    
         Open, Motion Alarm On\\
         
         Open, Change Brightness\\

         ...\\
    
         Open, Siren On \\
         Open, Alarm Action\\
         ...\\
    
         Person Detected, Power Off \\
    
         Smoke Detected, Power Off \\
        
         Doorbell Pressed, Power On\\
    \hline
    \end{tabular}
\end{adjustbox}
\captionof{table}{Wyze Smart Home Rules Dataset entities and trigger-action pairs.} \label{table:wyzedata}
\centering
\begin{adjustbox}{width=\columnwidth,center}
\centering

\begin{tabular}{|l|}
    \hline
         \textbf{Types of Entities}\\
    \hline
         Android Device\\
        Weather\\
        Gmail\\
        YouTube\\
        ...\\
        Facebook\\
        Instagram\\
        Linkedin\\
        ...\\
        Twitter\\
        Reddit\\
    \hline
    \end{tabular}
\centering
\begin{tabular}{|l|}
    \hline
         \textbf{Types of Trigger-Action Pairs}\\
    \hline
         New Post, Share a Link \\
        New Follower, Post a Tweet \\
        New Like, Add File from Url \\
        New Liked Video, Create a Post \\
         ... \\
        New Photo, Send Me an Email \\
        New Photo, Add File from Url \\
        New Screenshot, Add File from Url \\
        ...\\
        You Exit an Area, Set Temperature \\
        Battery Low, Send an Sms \\
    \hline
    \end{tabular}
\end{adjustbox}
\captionof{table}{IFTTT Dataset entities and trigger-action pairs.}
\label{table:iftttdata}
\end{table}

\paragraph{Evaluation Metrics} We use the following metrics to compare algorithms:
\begin{itemize}[nosep]
    \item Loss: Binary cross entropy loss with the positive and negative edges.
    \item AUC: Area Under the Curve.
    \item Mean Rank: Mean rank of positive testing edges between specific entities.
    \item Hit Rate@N: Recommend N rules and check if the positive test edges are included.
\end{itemize}

\subsection{Comparison of Graph-Based and User-Item-Based Methods}

\paragraph{Quantitative Evaluation} 
 \begin{figure}[ht]
    \centering
        \includegraphics[width = 0.43\textwidth]{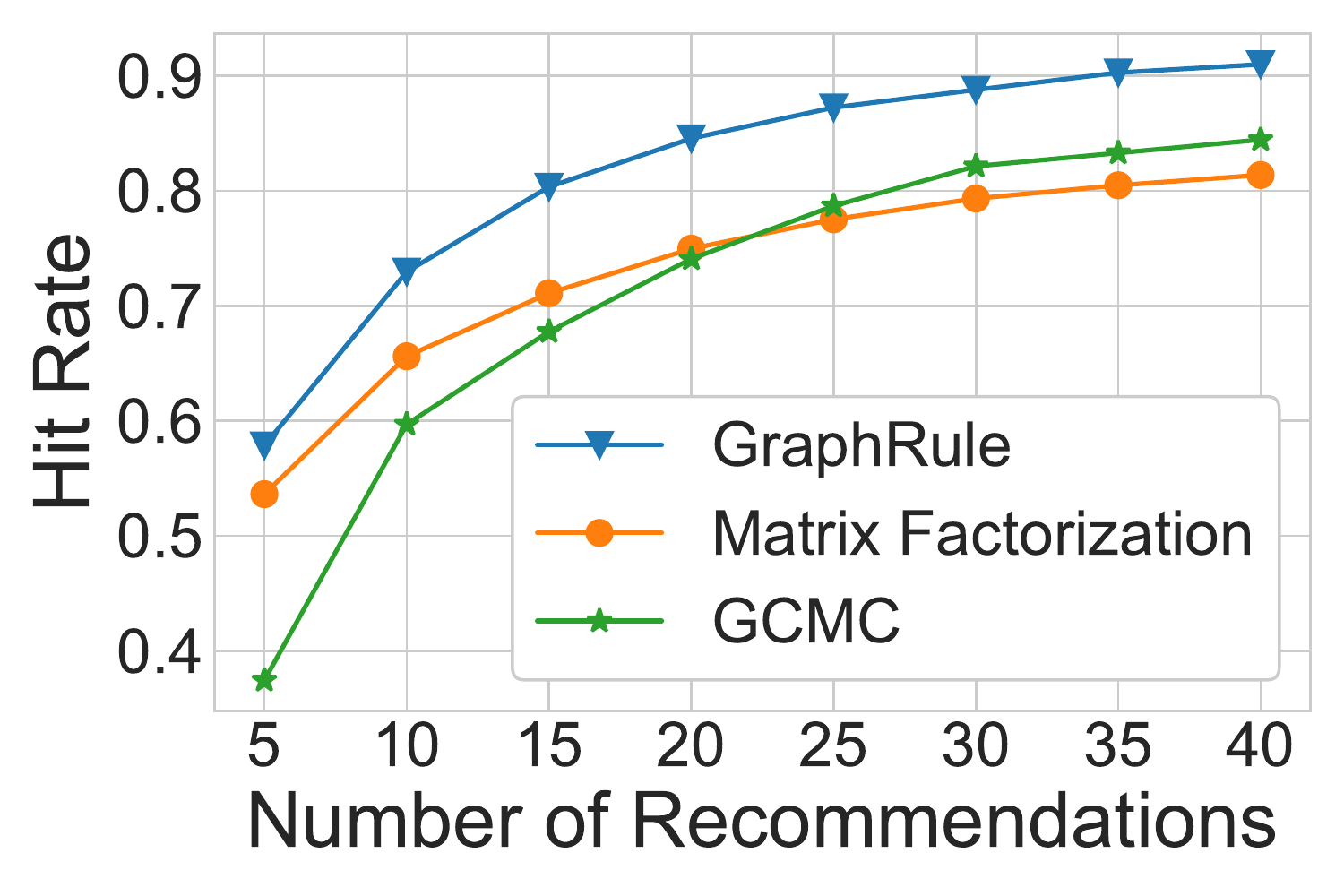}
    \caption{Hit Rate on the test set of Wyze Smart Home Rules Dataset vs. number of recommendations, comparing \graphrule with conventional recommendation systems. Recommendations are filtered by valid rules for a fair comparison.} 
    \label{fig:test_hit_rate_wyze_no_same_mf_gcmc}
\end{figure}

\begin{table}[ht]
    \centering
    \resizebox{1.0\columnwidth}{!}{
    \begin{tabular}{cccc}
    \toprule
                  \multicolumn{2}{c}{Triggers}  &  \multicolumn{2}{c}{Actions}
        \\ \cmidrule(r){1-2} \cmidrule(r){3-4}
                 Entity      & Trigger &  Action & Entity
        \\
        \midrule
        Contact Sensor & Open & Power On & Camera\\
        Contact Sensor & Open & Motion Alarm & Camera \\
        Contact Sensor & Open & Power On & Plug\\
        Contact Sensor & Open & Power On & Light
        \\ \midrule
        \rowcolor{LightCyan}
        Contact Sensor & Open & Notifications & Camera\\
        \bottomrule
    \end{tabular}
    }
\caption{Arbitrary user's rules. The last rule is omitted from the users' rules for prediction.}
\label{table:exp_rules}
\end{table}

\begin{table*}[ht]
    \centering
    \resizebox{\textwidth}{!}{
    \begin{tabular}{lllllll}
    \toprule
                 & \multicolumn{2}{c}{\graphrule}  &  \multicolumn{2}{c}{Matrix Factorization} & \multicolumn{2}{c}{GCMC}
        \\ \cmidrule(r){2-3} \cmidrule(r){4-5} \cmidrule(r){6-7}
                 Rank      & Recommendation Rule &  Prob.   & Recommendation Rule &  Prob.    &  Recommendation Rule &  Prob.
        \\
        \midrule
        1 & (Contact Sensor, Open, Change Brightness, Light) & 0.997 
        & (Camera, Motion Detected, Alarm, Camera) & 0.999
        & (Motion Sensor, Motion Detected, Power On, Light) & 0.910\\
         2 & (Contact Sensor, Open, Alarm, Camera) & 0.996 
        & (Camera, Motion Detected, Motion Alarm, Camera) & 0.980
        & (Motion Sensor, Motion Detected, Change Brightness, Light) & 0.855\\
         3 & \mycc{(Contact Sensor, Open, Notifications, Camera)} & \mycc{0.995} 
        & (Camera, Motion Detected, Notification, Camera) & 0.972
        & (Motion Sensor, Being Off, Power Off, Light) & 0.840\\
        & \multicolumn{2}{c}{$\vdots$} &  \multicolumn{2}{c}{$\vdots$} & \multicolumn{2}{c}{$\vdots$}\\
        17 & (Contact Sensor, Being Open, Notification On, Camera) & 0.882 
        & (Camera, Sound Detected, Notification, Cloud) & 0.335
        & \mycc{(Contact Sensor, Open, Notifications, Camera)} & \mycc{0.623}\\
        18 & (Contact Sensor, Condition Met, Alarm, Camera) & 0.860
        & (Camera, Condition Met, Power On, Camera) & 0.322
        & (Motion Sensor, Condition Met, Power On, Light) & 0.592\\
        19 & (Contact Sensor,Being Open , Power On, Camera) & 0.860 
        & (Camera, Condition Met, Motion Alarm, Camera) & 0.321
        & (Camera, Motion Detected, Alarm, Camera) & 0.552\\
        20 & (Contact Sensor, Being Open, Alarm, Camera) & 0.851 
        & (Camera, Condition Met, Notification, Camera) & 0.315
        & (Camera, Person Detected, Notification, Cloud) & 0.541\\
        \bottomrule
    \end{tabular}
    }
\caption{Qualitative comparison of \graphrule, Matrix Factorization, and GCMC methods in predicting the removed rule from the rule set of an arbitrary user in Table~\ref{table:exp_rules}. The removed rule is the third recommendation in \graphrule, but it is the $17$th one in GCMC. The removed rule is not included in the first 20 rules recommended by matrix factorization.}
\label{table:exp_recom}
\end{table*}

We first compare our centralized graph model, \graphrule, with user-item-based methods. For a fair comparison with user-item-based methods, we consider entities with the same type as one entity. As shown in Figure~\ref{fig:test_hit_rate_wyze_no_same_mf_gcmc}, the hit rate of \graphrule is outperforming both GCMC and Matrix Factorization in this task, which validates our analysis in Section \ref{sec:rule_recommendation}. With the increase of the number of recommendations, \graphrule has a better hit rate, close to $0.91$, when recommends $40$ rules, which means by recommending $40$ rules among $1207$ rules total, the user has $91\%$ chance to choose rules in the recommendation lists on average.

\paragraph{Qualitative Evaluation}
To qualitatively evaluate the performance of these algorithms we provide a case, where we feed a user's rule data to each of the models while omitting one of the rules set by that user. The user's rule examples are described in Table~\ref{table:exp_rules}. Then, we show what is the rank of that omitted rule in the recommendations made by each algorithm as well as other top recommendations in Table~\ref{table:exp_recom}. As it can be seen in Table~\ref{table:exp_recom}, the removed rule is the third rank in the recommendation by \graphrule, while it is the $17$th by GCMC. It was not in any of the top $20$ recommendations of the matrix factorization approach. Also, note that most of the recommendations by matrix factorization are for cameras, which is due to the popularity of camera rules among other users. Also, GCMC top rules are for a motion sensor that is not available in the user's entity set. On the other hand, all recommendations by \graphrule are applicable to the current set of entities for the user.

\subsection{Evaluation of Centralized and Federated Algorithms}
In this part, we evaluate the performance of our proposed \fedrule, and compare it with FedGNN training, as well as our proposed centralized \graphrule. We perform the evaluations on both train and test datasets.

\begin{figure}[t]
    \centering
    \includegraphics[width = 0.23\textwidth]{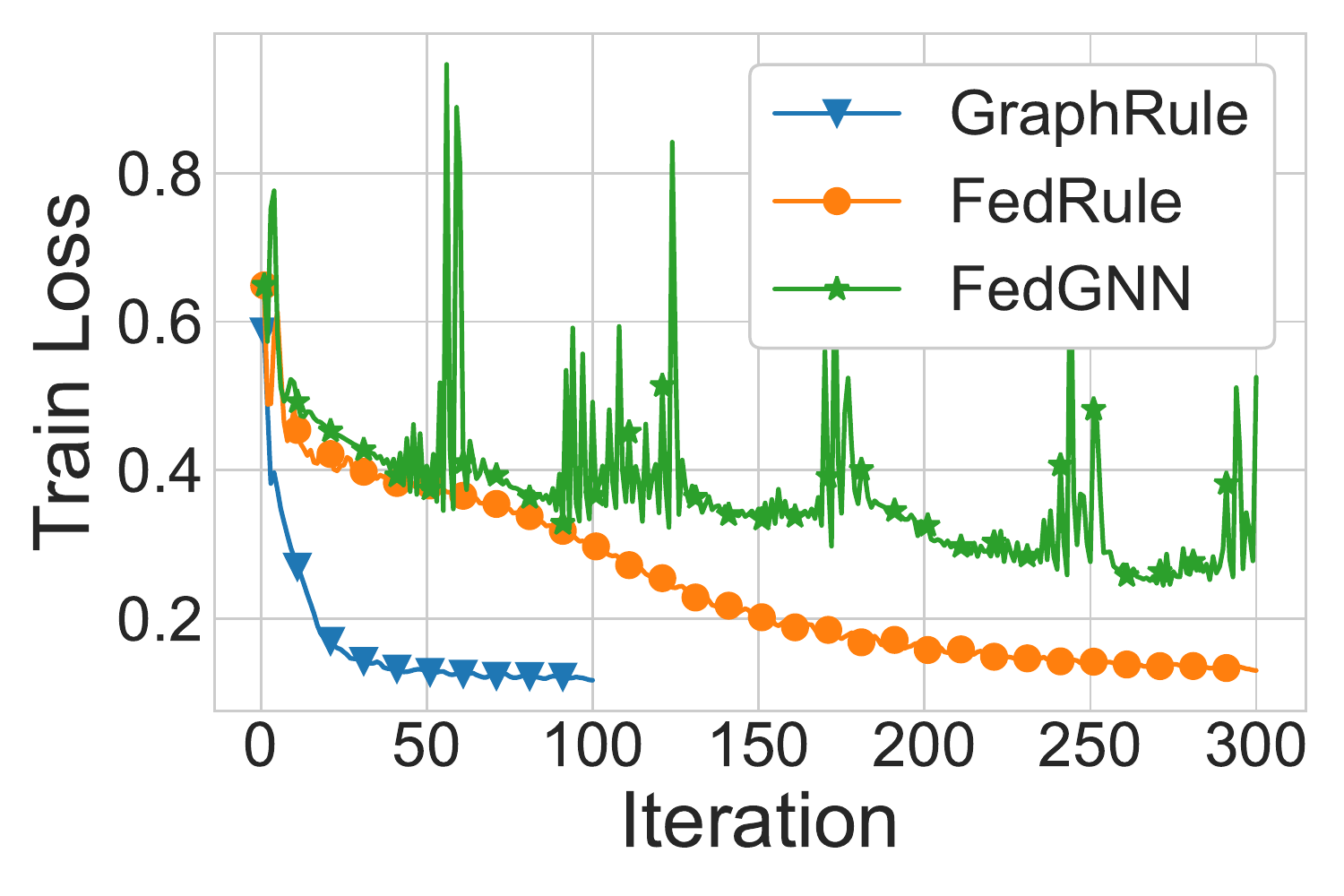}
    \includegraphics[width = 0.23\textwidth]{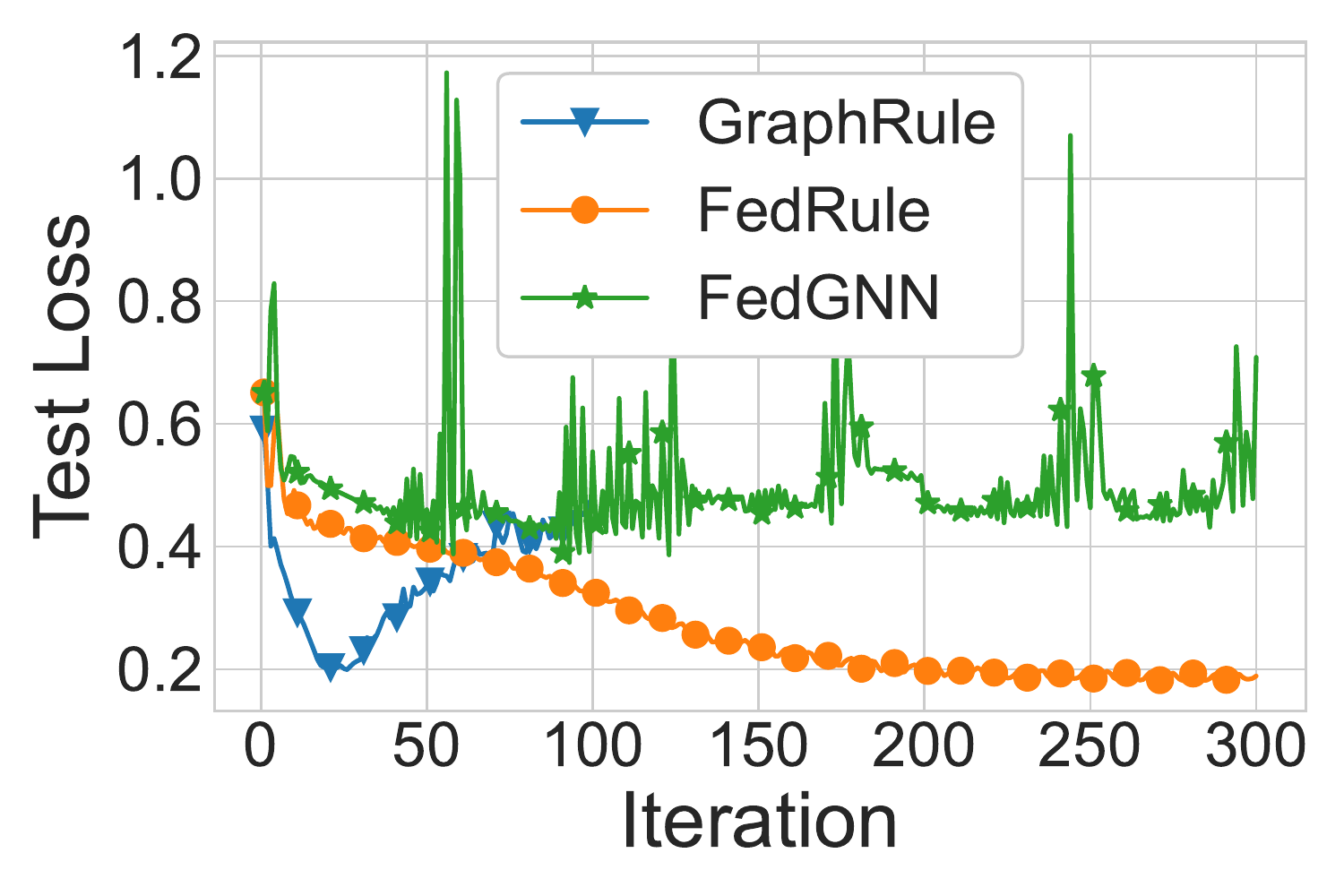}
    \includegraphics[width = 0.23\textwidth]{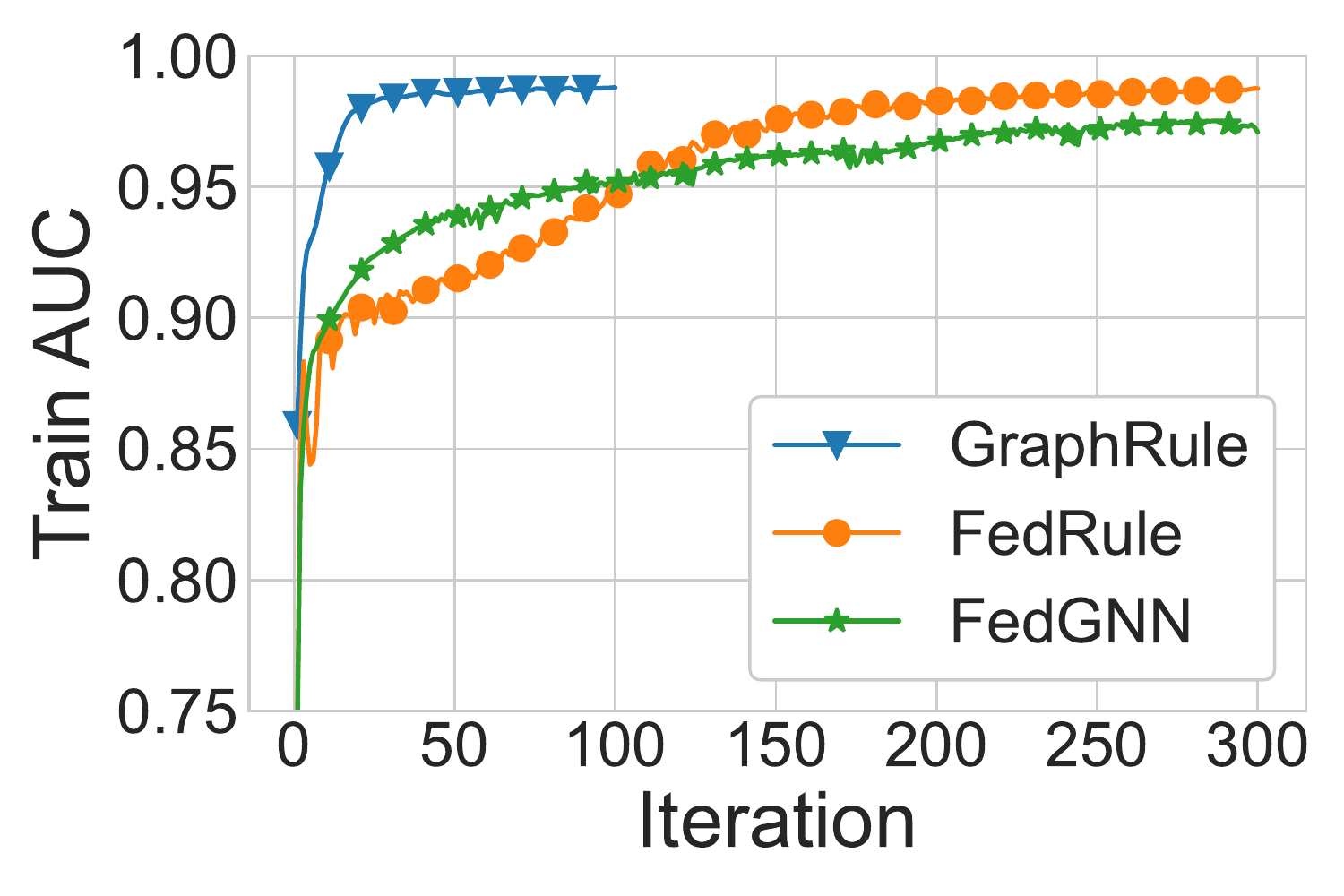}
    \includegraphics[width = 0.23\textwidth]{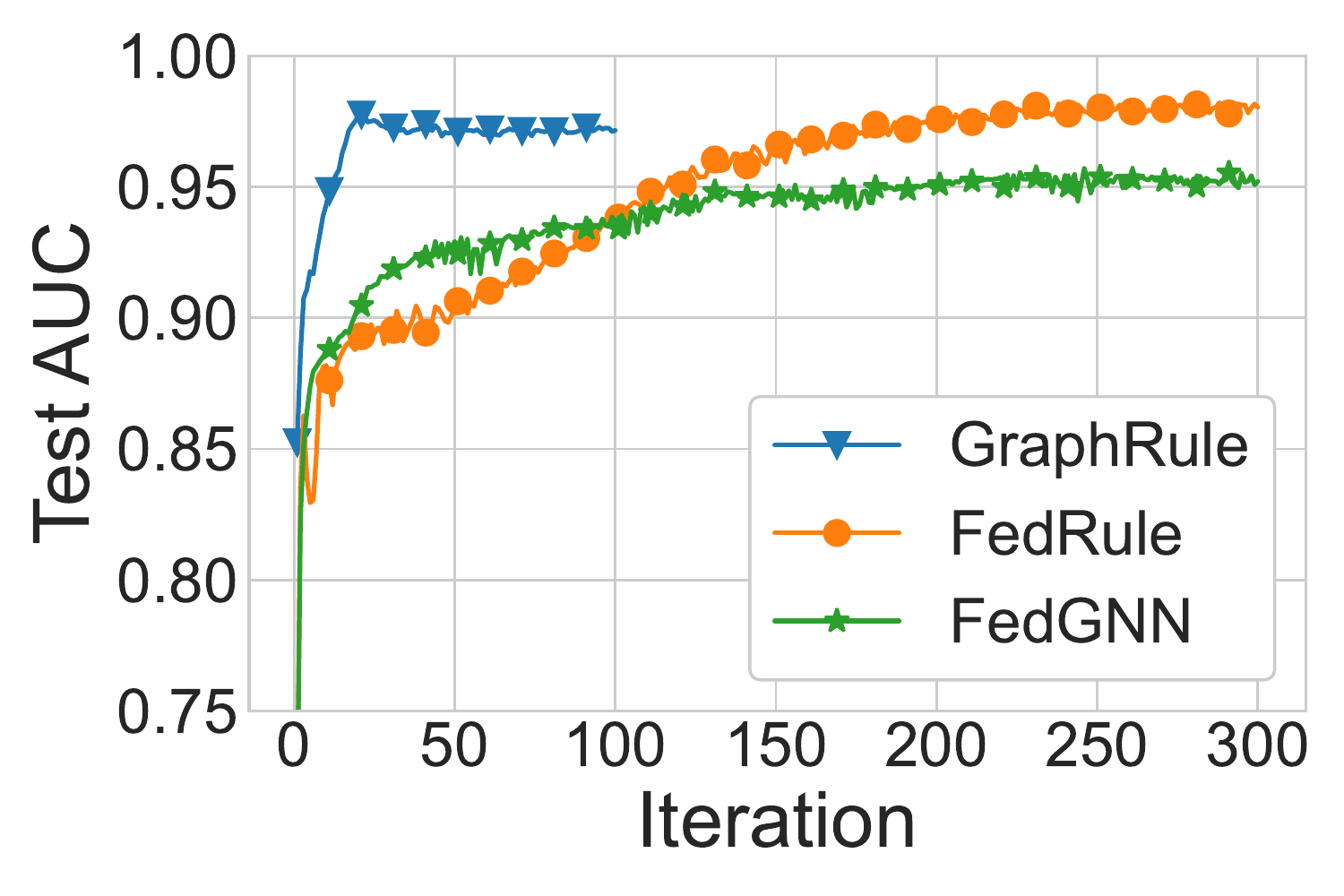}
    \includegraphics[width = 0.23\textwidth]{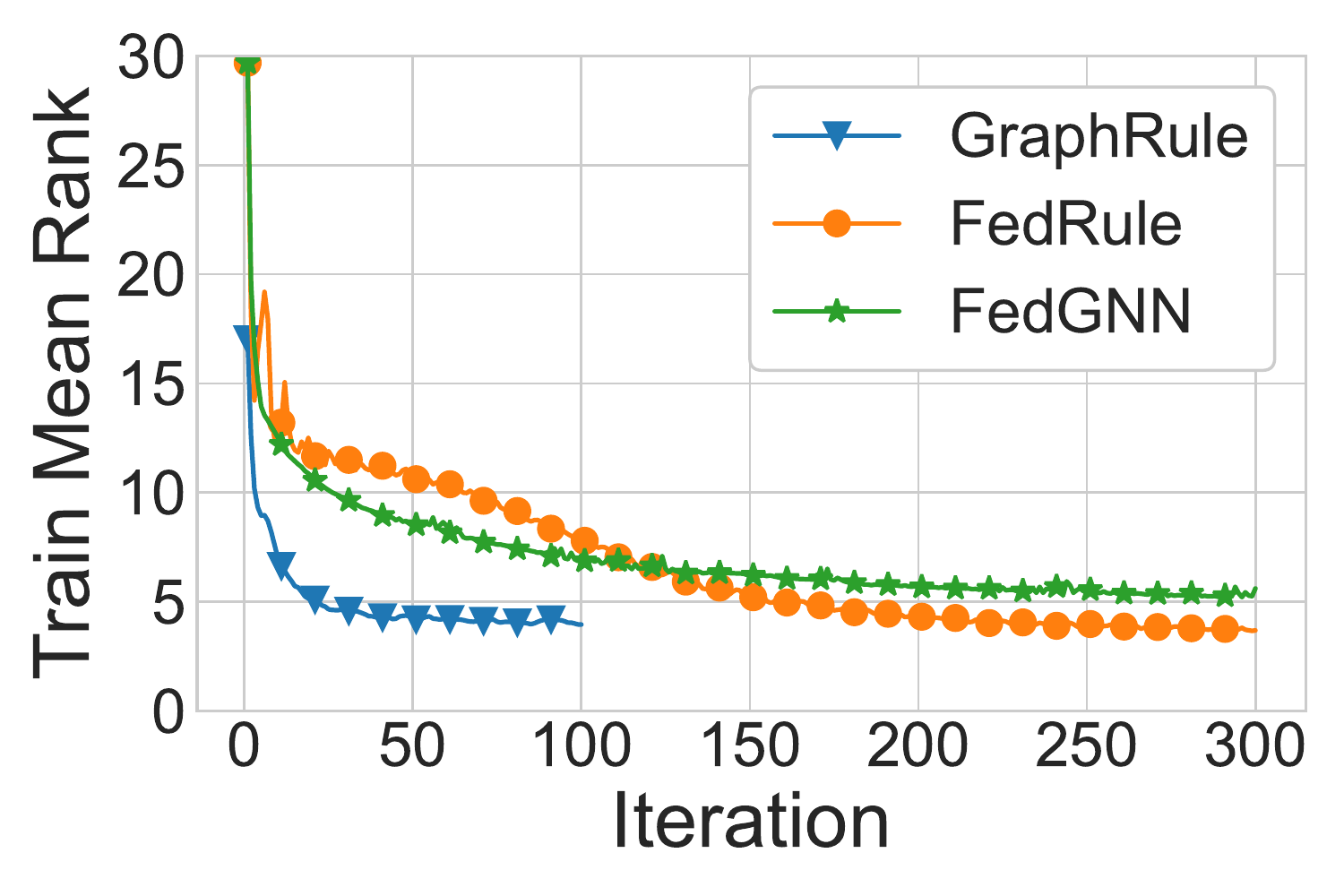}
    \includegraphics[width = 0.23\textwidth]{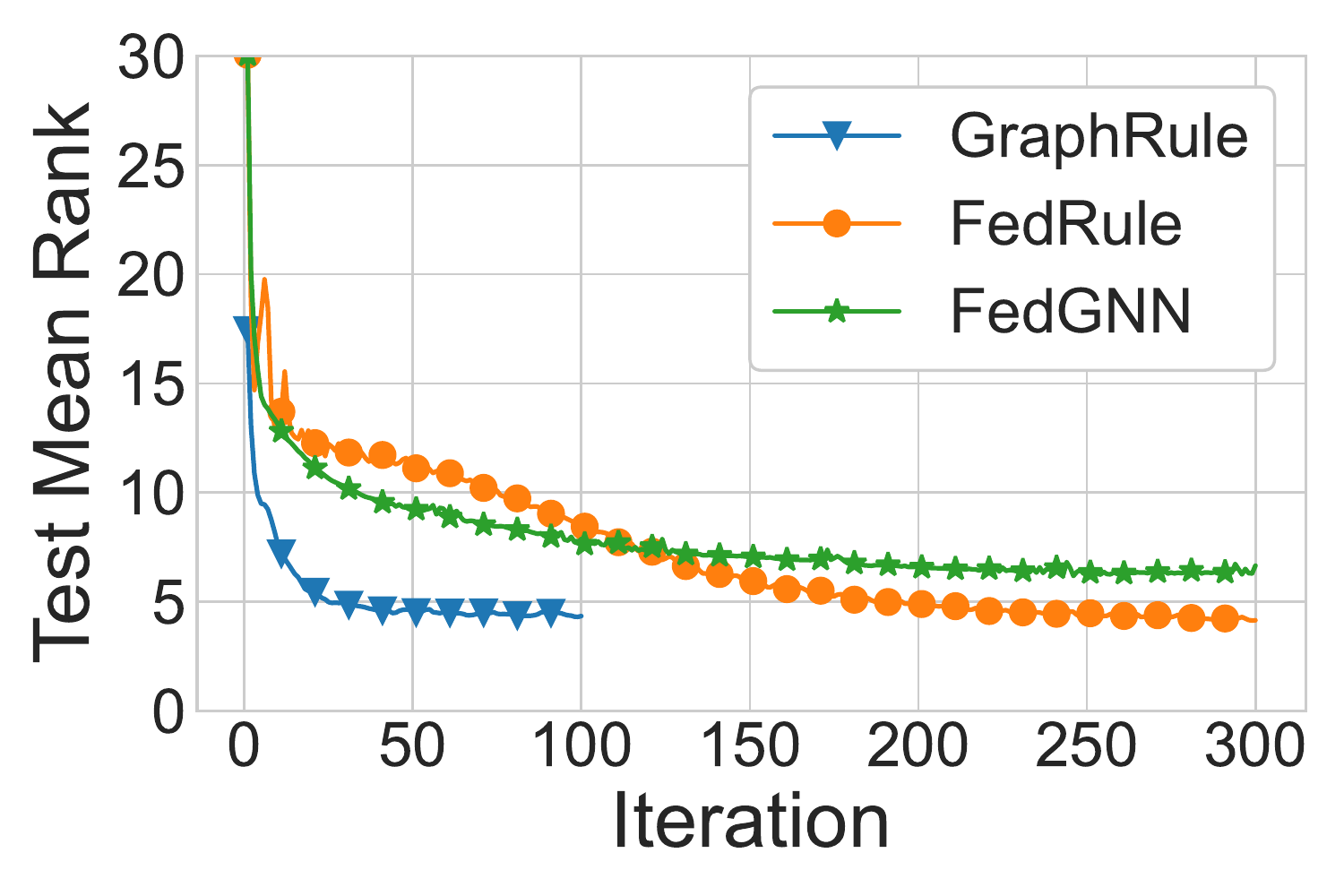}
    \caption{Train and test performance of different algorithms on Wyze Smart Home Rules Dataset. \fedrule smoothly converges while FedGNN training diverges after a while.}
    \label{fig:train_result_wyze_no_same}
\end{figure}

\begin{table}[t]
  {%
\centering
  \begin{tabular}{|l|l|l|l|l|l|}
\hline
 &
  Loss &
  AUC &
  MR &
  MR(RT)  \\ \hline
\graphrule & 0.1997          & 0.9768        & 4.349  & 3.0970    \\ \hline
FedGNN      & 0.3878          & 0.9521       & 6.661    & 5.3946             \\ \hline
\fedrule     & \textbf{0.1892} & \textbf{0.9804}  & \textbf{4.156} & \textbf{2.9150}          \\ \hline
\end{tabular}}
 \captionof{table}{Final test results on Wyze Smart Home Rules Dataset (RT: remove rules shown in training graph).}
\label{fig:test_table_company}

\end{table}

\begin{figure}[t]
    \centering
        \centering
        \includegraphics[width = 0.43\textwidth]{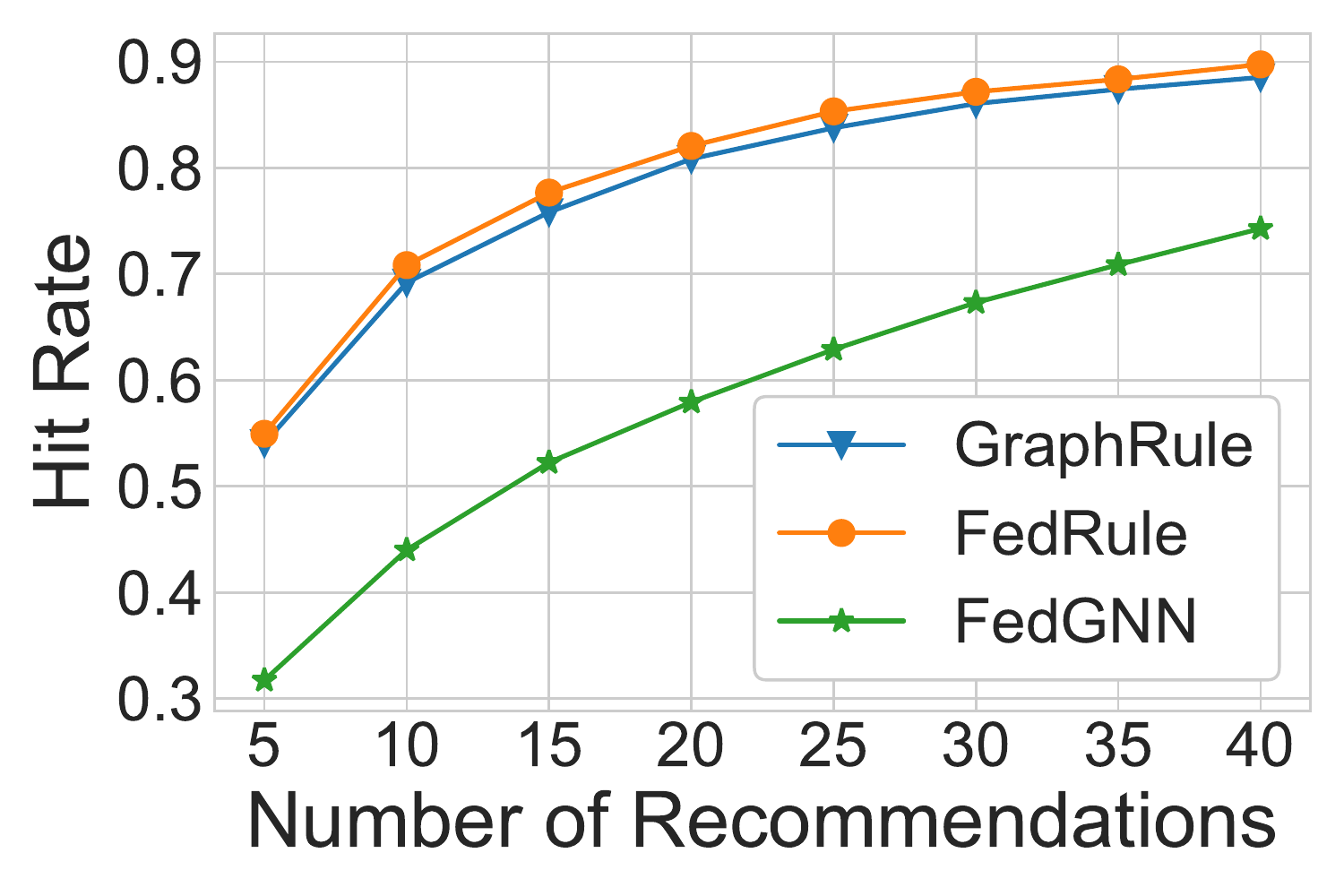}
  \caption{Hit Rate on the test set of Wyze Smart Home Rules Dataset vs. the number of recommendations, comparing \graphrule with federated approaches.}
    \label{fig:test_hit_rate_wyze_no_same}
\end{figure}

\paragraph{Results on Wyze Smart Home Rules Dataset}~The results of applying the aforementioned algorithms on the Wyze Smart Home Rules dataset are depicted in Figure~\ref{fig:train_result_wyze_no_same}. During the training, the centralized algorithm, \graphrule, converges faster since it utilizes the training data of all users in an IID manner. For federated settings, given the heterogeneous data distribution among users and the heterogeneity of the number of rules and entities for each user, FedGNN diverges after 50 iterations. This is especially exacerbated by a large number of users with a small amount of data (76, 218 users and 2.65 rules per user on average). On the other hand, \fedrule smoothly converges to the same loss as the \graphrule due to its variance reduction mechanism. The test loss shows that even \graphrule overfits after $20$ iterations and FedGNN diverges after 50 iterations. The test loss of \fedrule smoothly decreases and converges to the same value as the minimum loss in \graphrule after $200$ iteration. Similarly, the test AUC of \graphrule converges faster. During the training process, FedGNN has a better test AUC at the start but increases slowly after $50$ iterations as the training loss diverges. The test AUC of \fedrule increases steadily to the same level as \graphrule after $200$ iteration. The test Mean Rank also shows similar patterns.

Table~\ref{fig:test_table_company} shows the final test results of Wyze Smart Home Rules data. Although the \graphrule has access to all users' graphs during the training, \fedrule slightly outperforms \graphrule since it employs variance reduction, making the convergence more smooth. \fedrule greatly outperforms FedGNN in all criteria. The mean rank of \fedrule is $2.915$ after removing the rules in training graphs, which means that for any rules between two specific entities, we need to recommend three rules on average and the user is very likely to adopt at least one of them. Also, Figure~\ref{fig:test_hit_rate_wyze_no_same} shows the test hit rate of the Wyze Smart Home Rules dataset. \fedrule and \graphrule have very close performance and are at most $26.8\%$ better than FedGNN.

\begin{figure}[t]
    \centering
    \includegraphics[width = 0.23\textwidth]{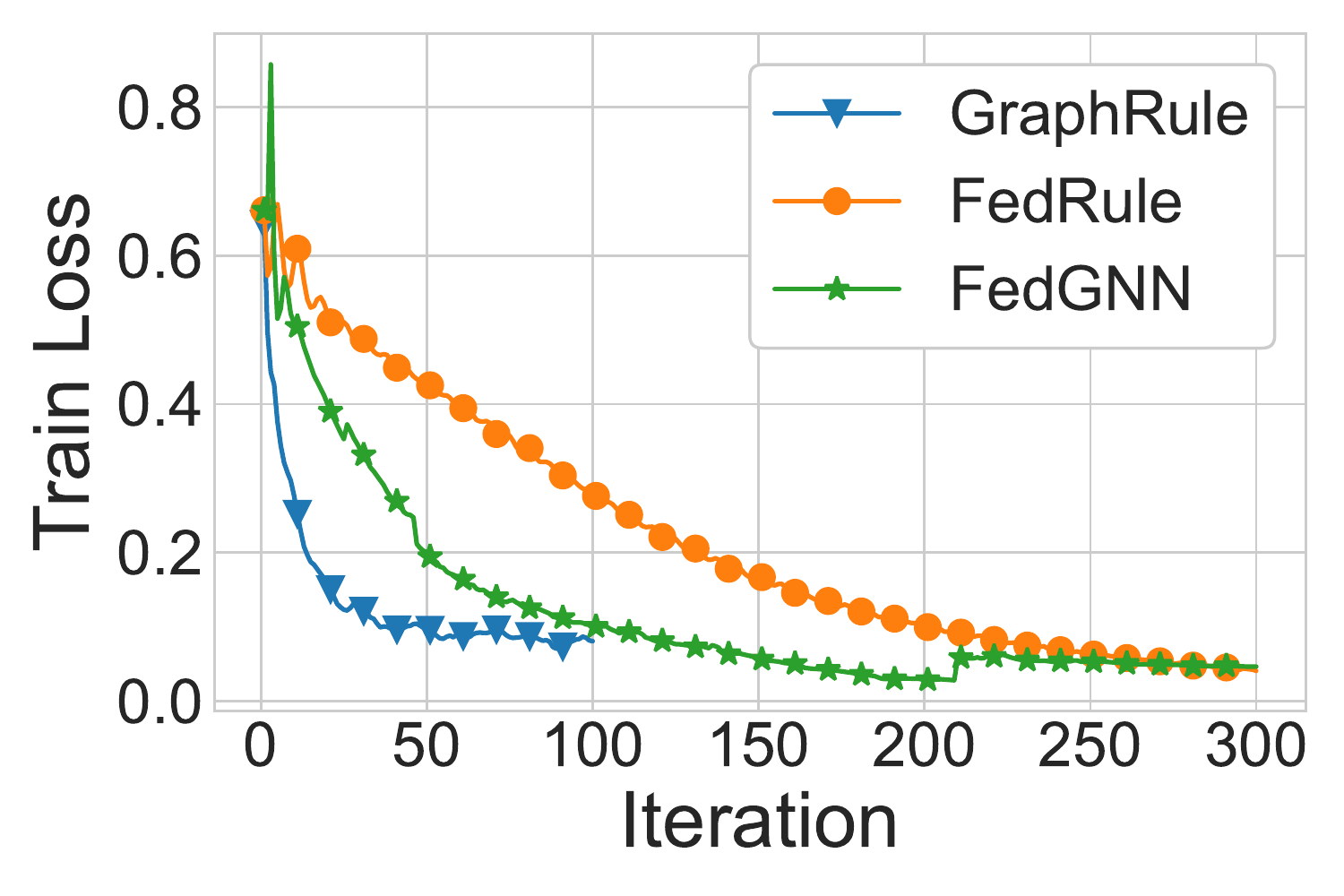}
    \includegraphics[width = 0.23\textwidth]{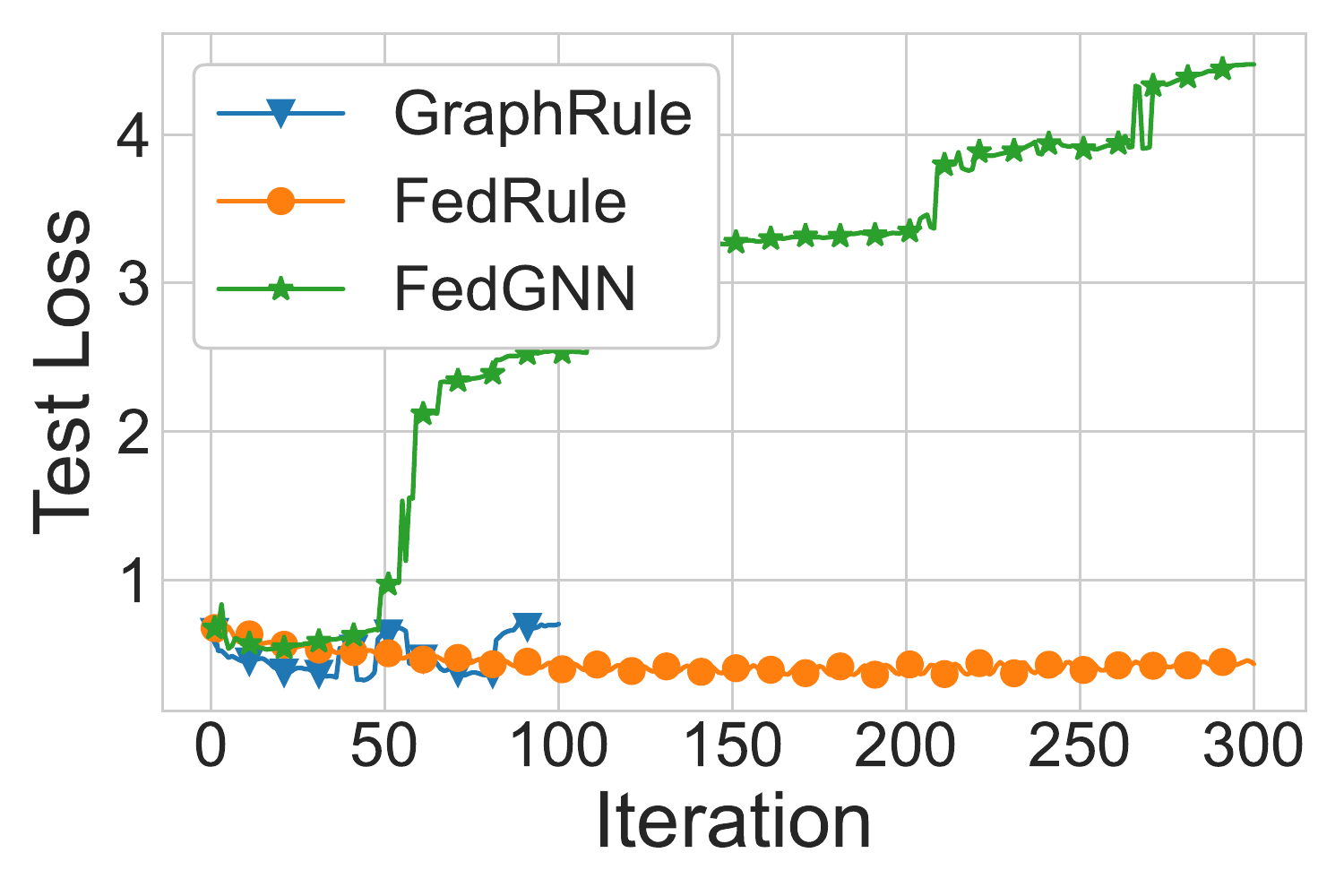}
    \includegraphics[width = 0.23\textwidth]{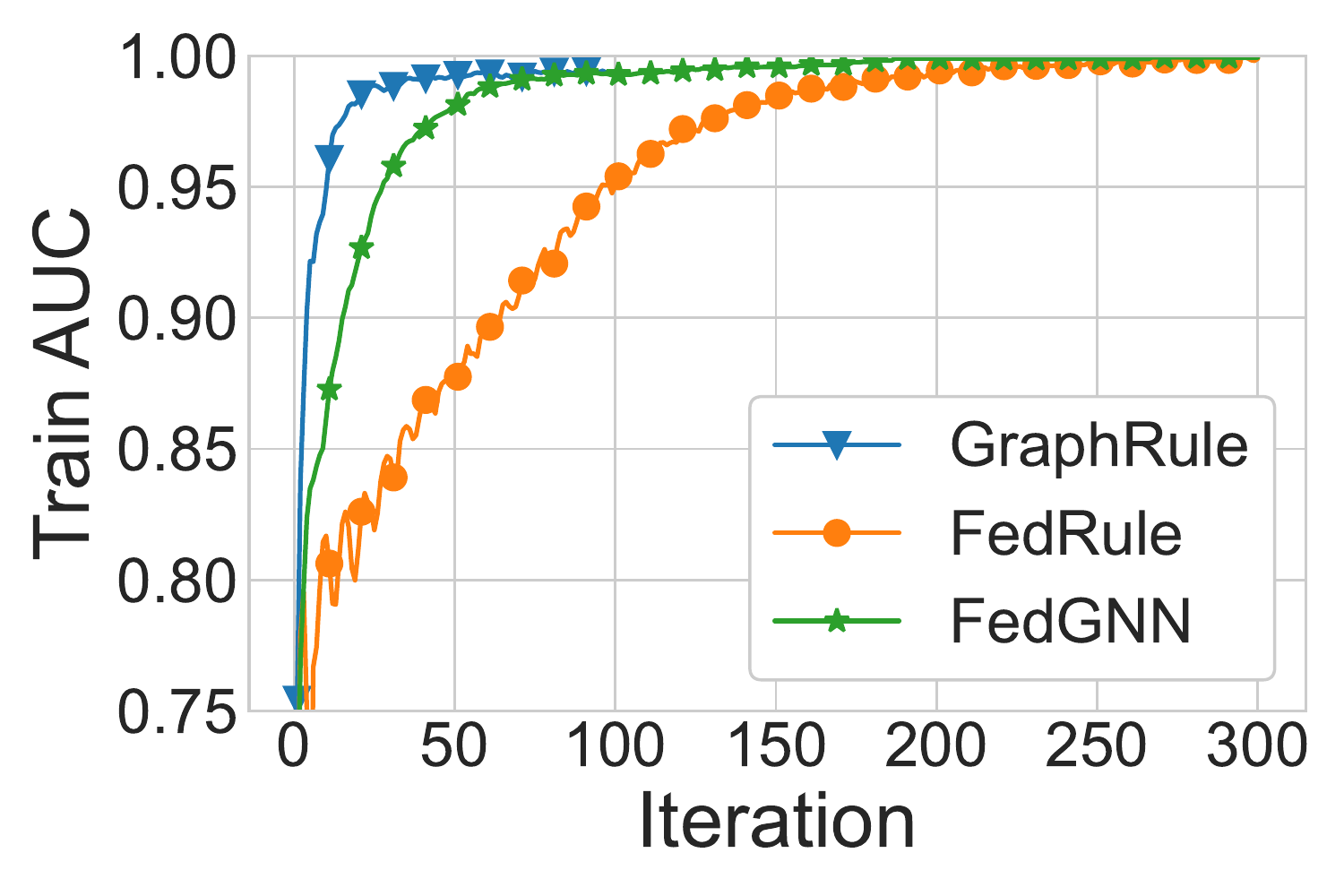}
    \includegraphics[width = 0.23\textwidth]{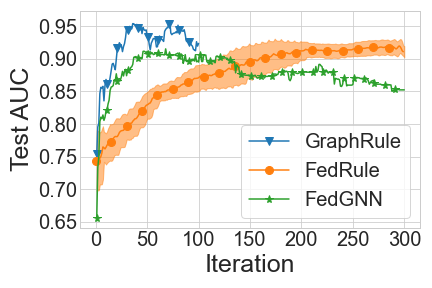} 
    \includegraphics[width = 0.23\textwidth]{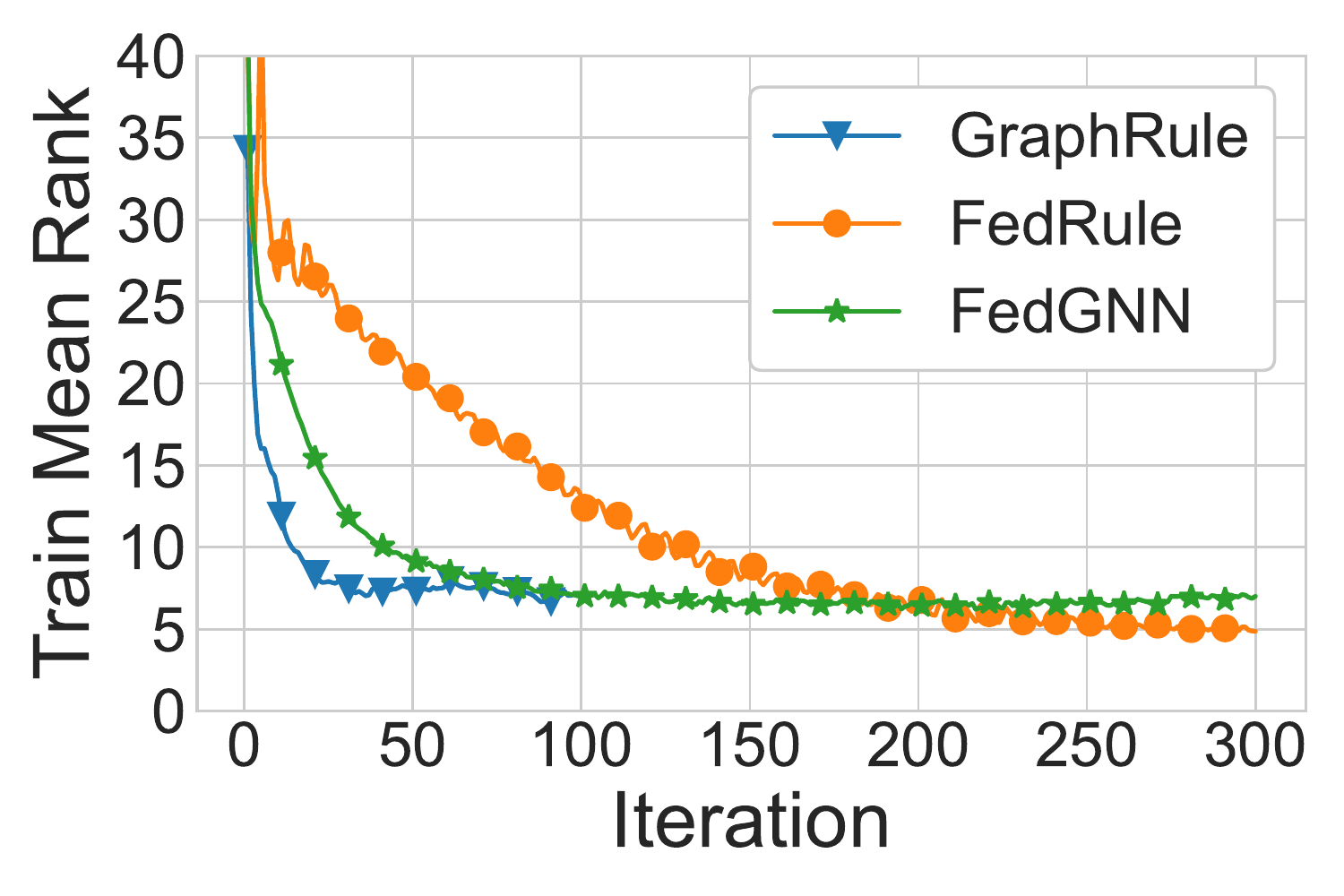}
    \includegraphics[width = 0.23\textwidth]{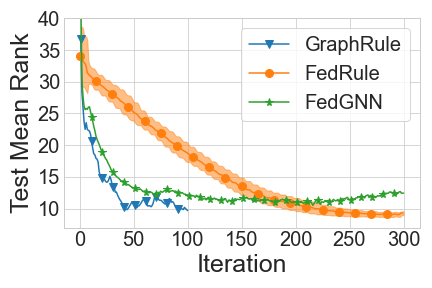}
    
    \caption{Training and test performance of different centralized and federated algorithms on IFTTT Dataset. Again, \fedrule smoothly converges while FedGNN overfits.}
    \label{fig:train_result_ifttt_no_same}
\end{figure}

\begin{table}[t]
    {%
\centering
\begin{tabular}{|l|l|l|l|l|l|}
\hline
 &
  Loss &
  AUC &
  MR &
  MR(RT)  \\ \hline
\graphrule & \textbf{0.3238} & \textbf{0.9491} & 9.6997  & 9.6745 \\ \hline
FedGNN     & 0.4905 & 0.9072 & 12.4479    & 12.4240 \\ \hline
\fedrule     & 0.3614 & 0.9417  & \textbf{8.8812} & \textbf{8.8564} \\ \hline
\end{tabular}}
\captionof{table}{Final test results on IFTTT Dataset. (RT: remove rules shown in training graph).}
\label{fig:test_table_ifttt}
\end{table}

\begin{figure}[t]
        \centering
        \includegraphics[width = 0.43\textwidth]{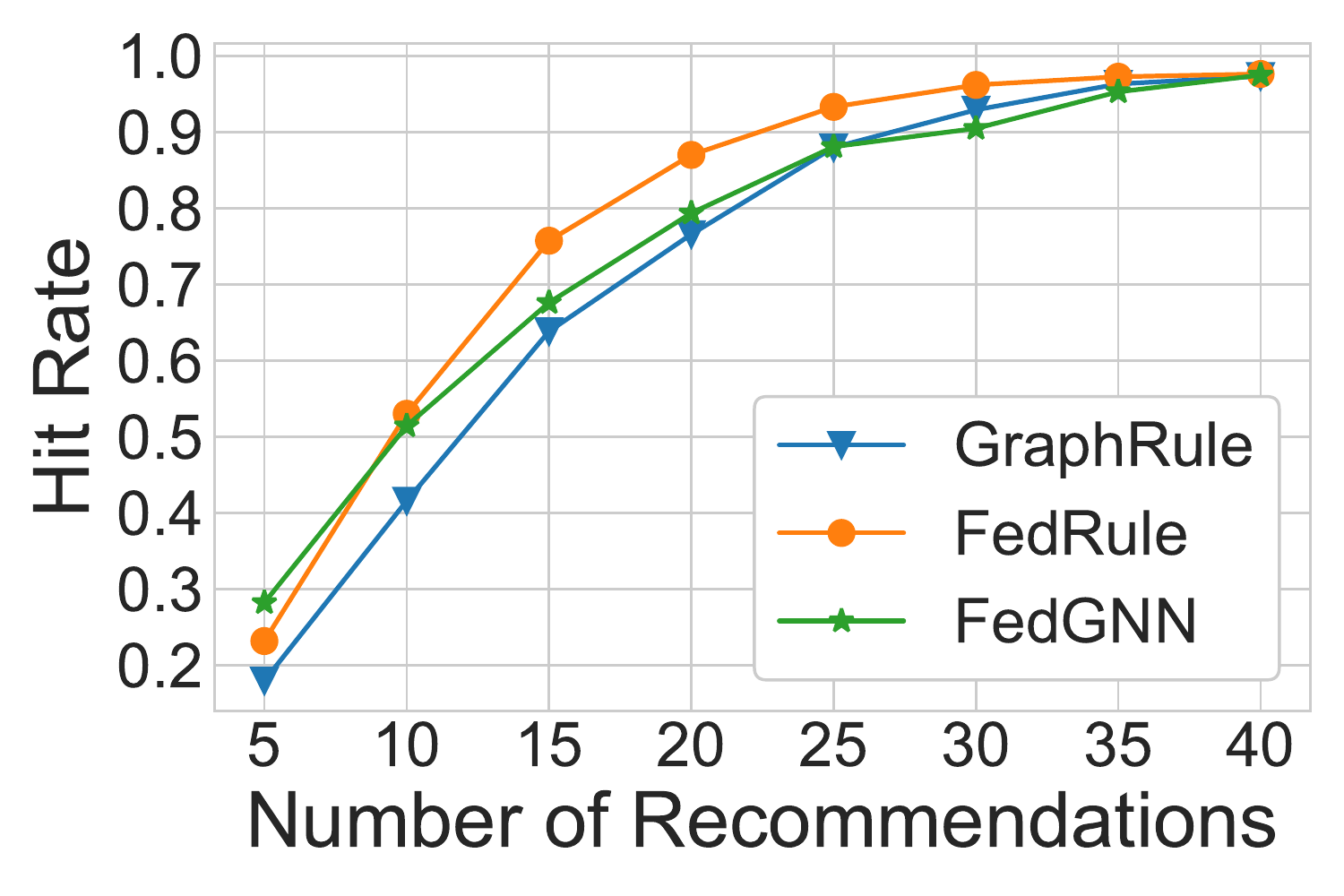}
    \caption{The Hit Rate on the IFTTT test dataset for centralized and federated algorithms with the rule filter.}
    \label{fig:test_hit_rate_ifttt_no_same}
\end{figure}

\begin{figure}[t]
    \centering
    \includegraphics[width = 0.23\textwidth]{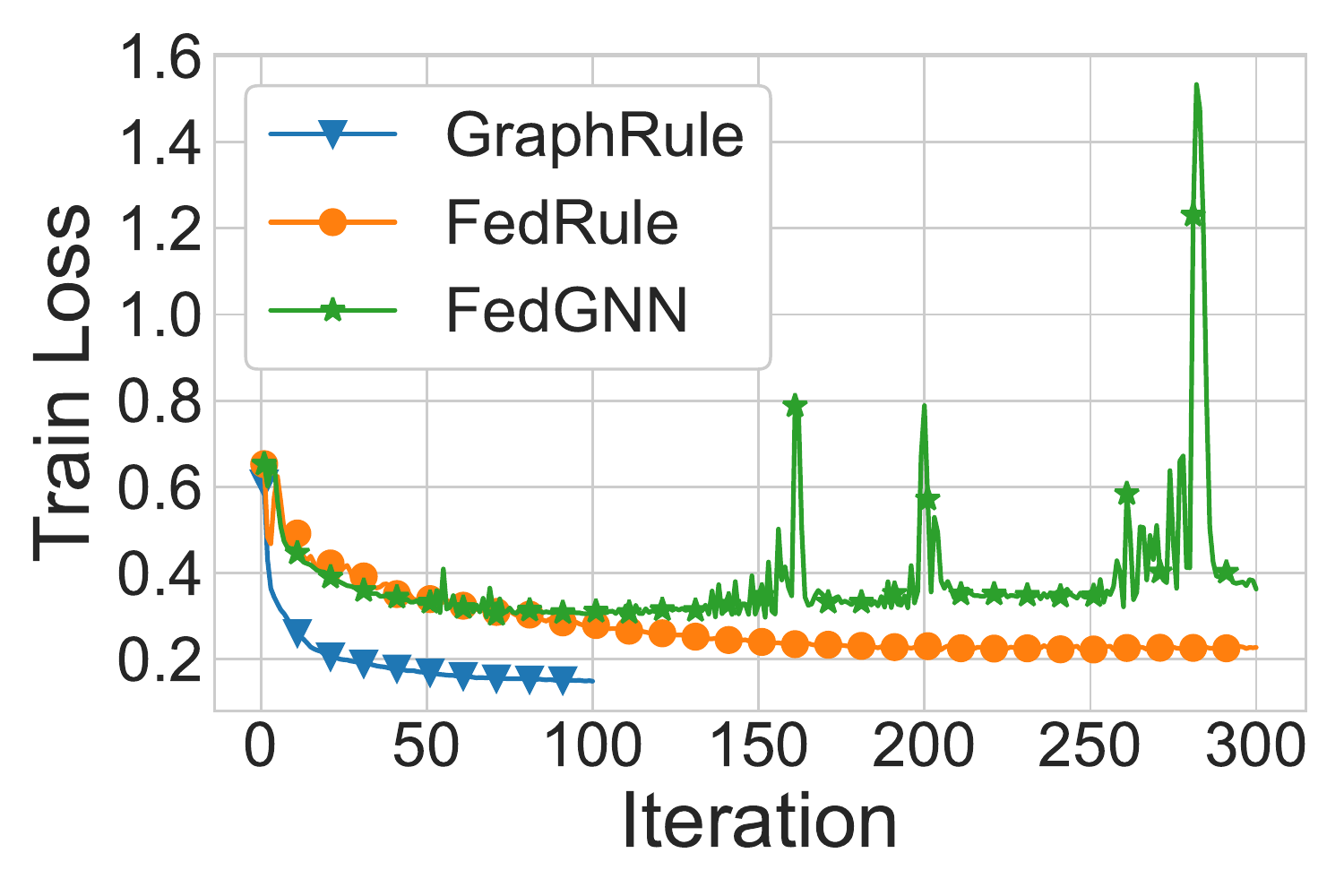}
    \includegraphics[width = 0.23\textwidth]{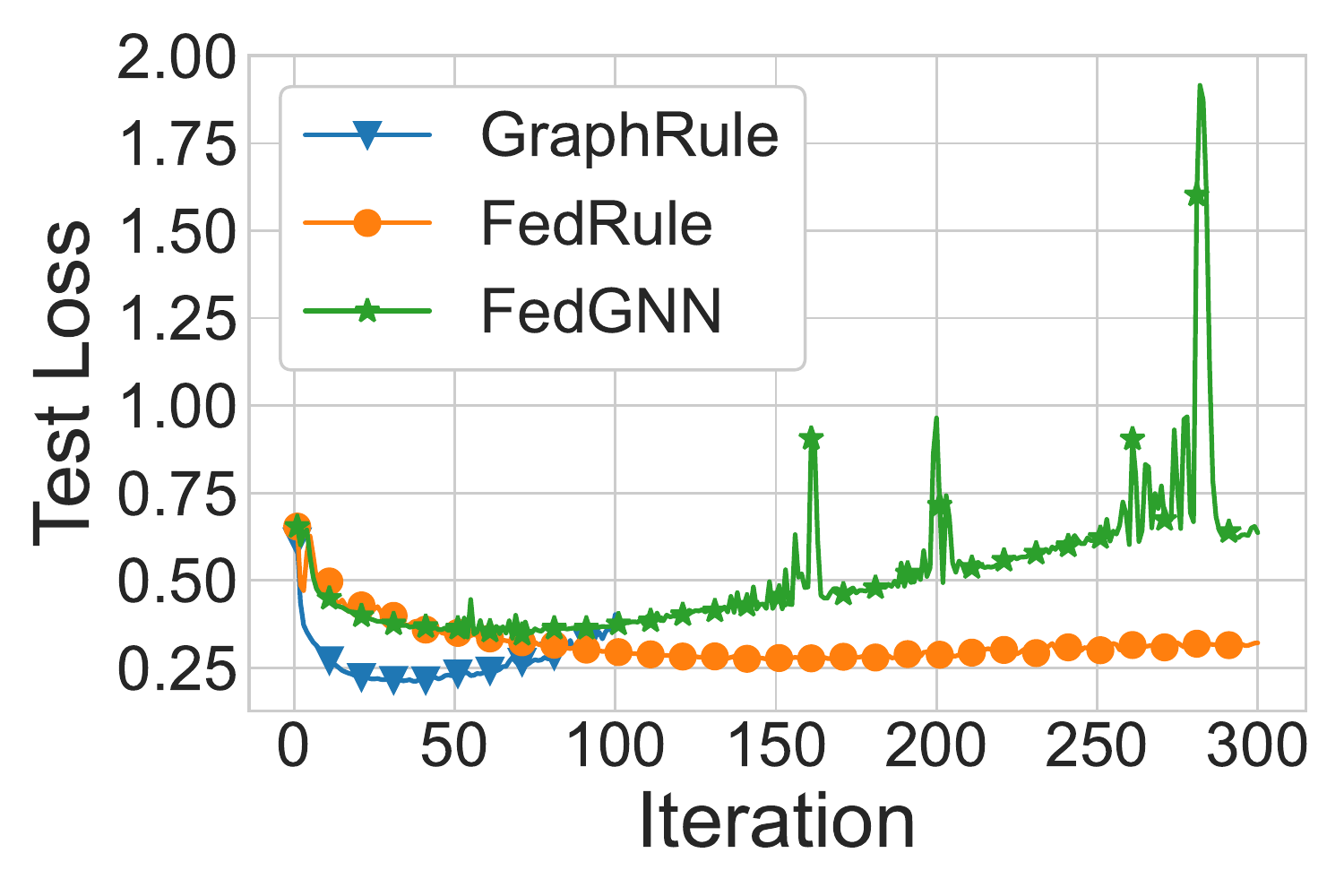}
    \includegraphics[width = 0.23\textwidth]{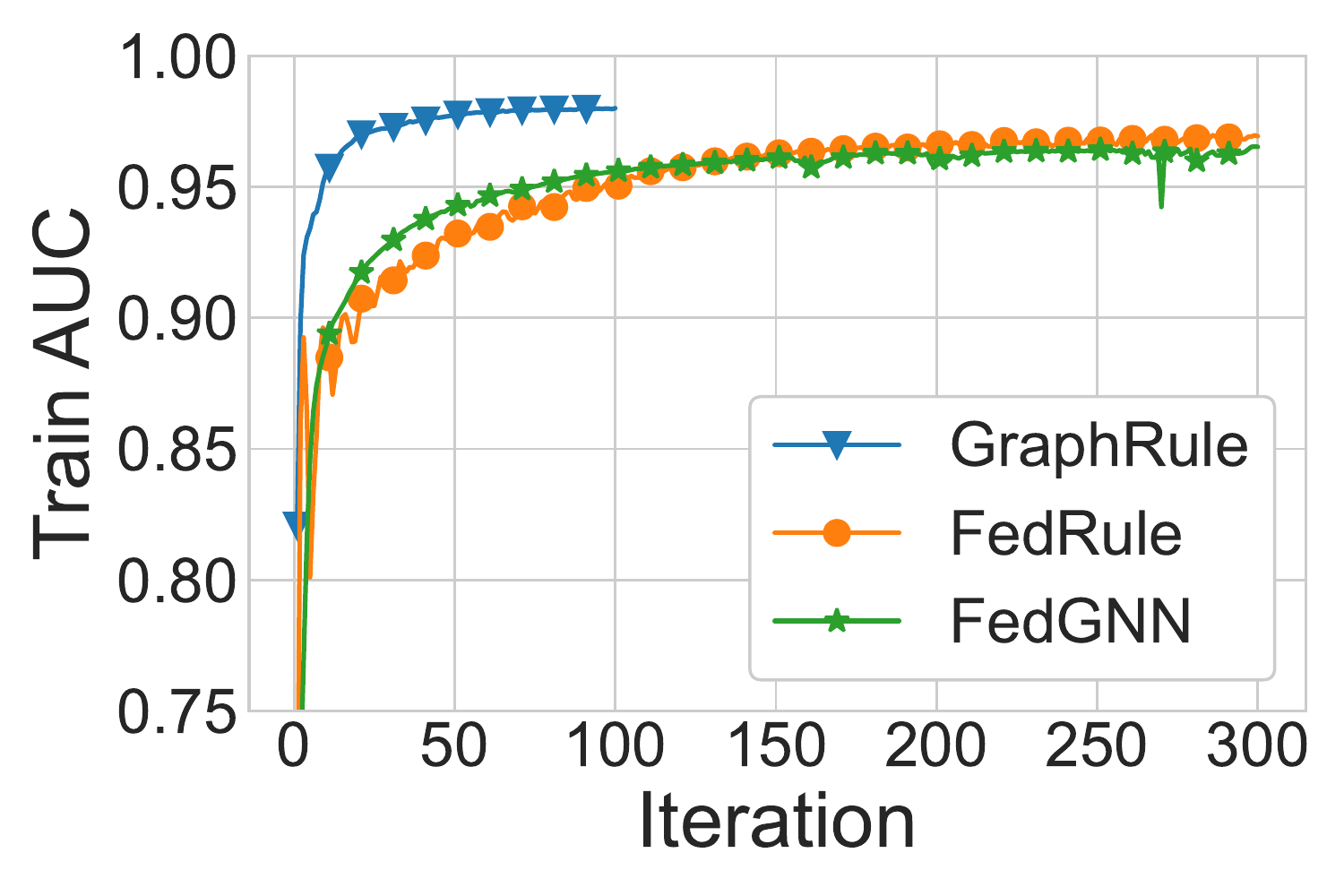}
    \includegraphics[width =
    0.23\textwidth]{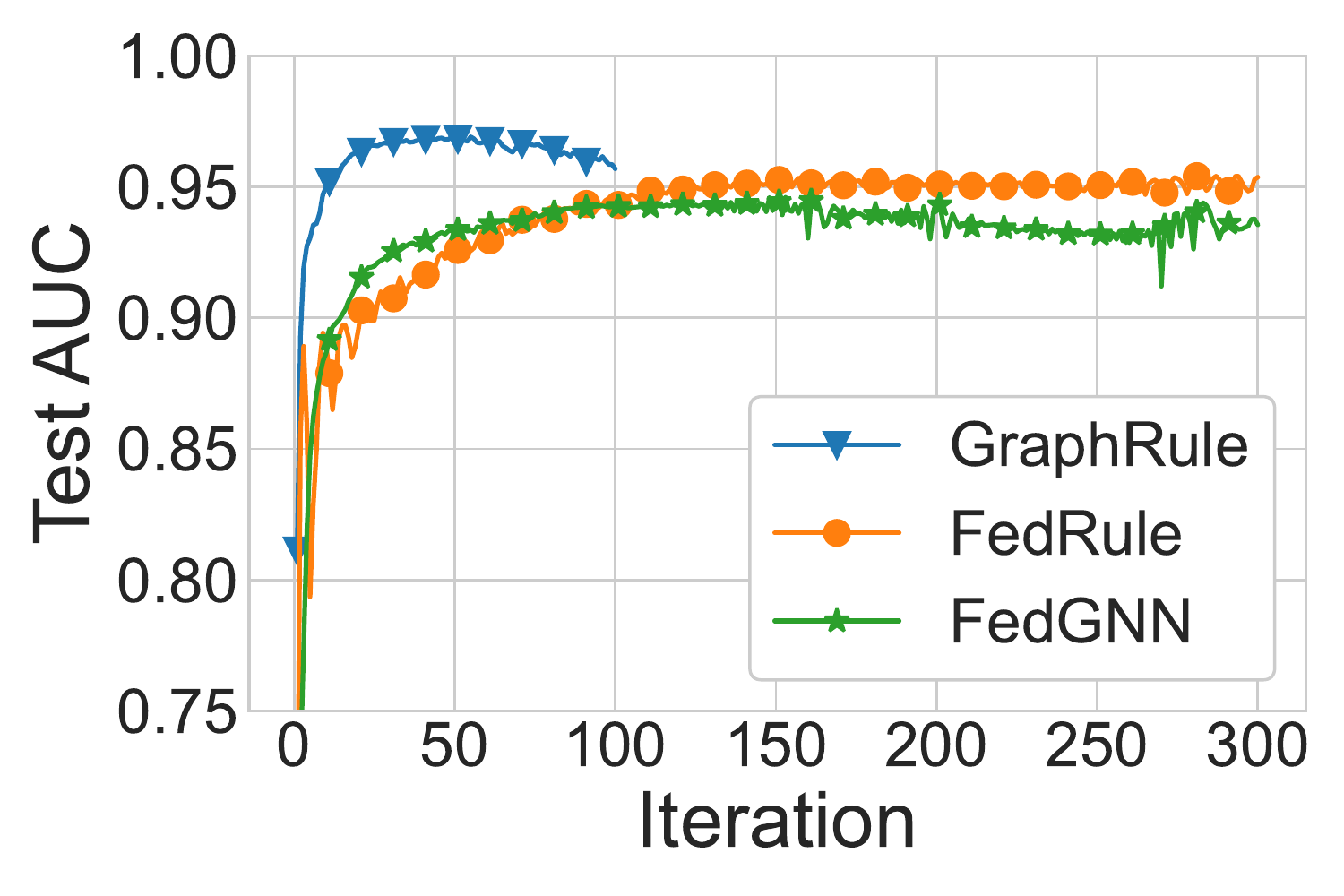}
    \includegraphics[width = 0.23\textwidth]{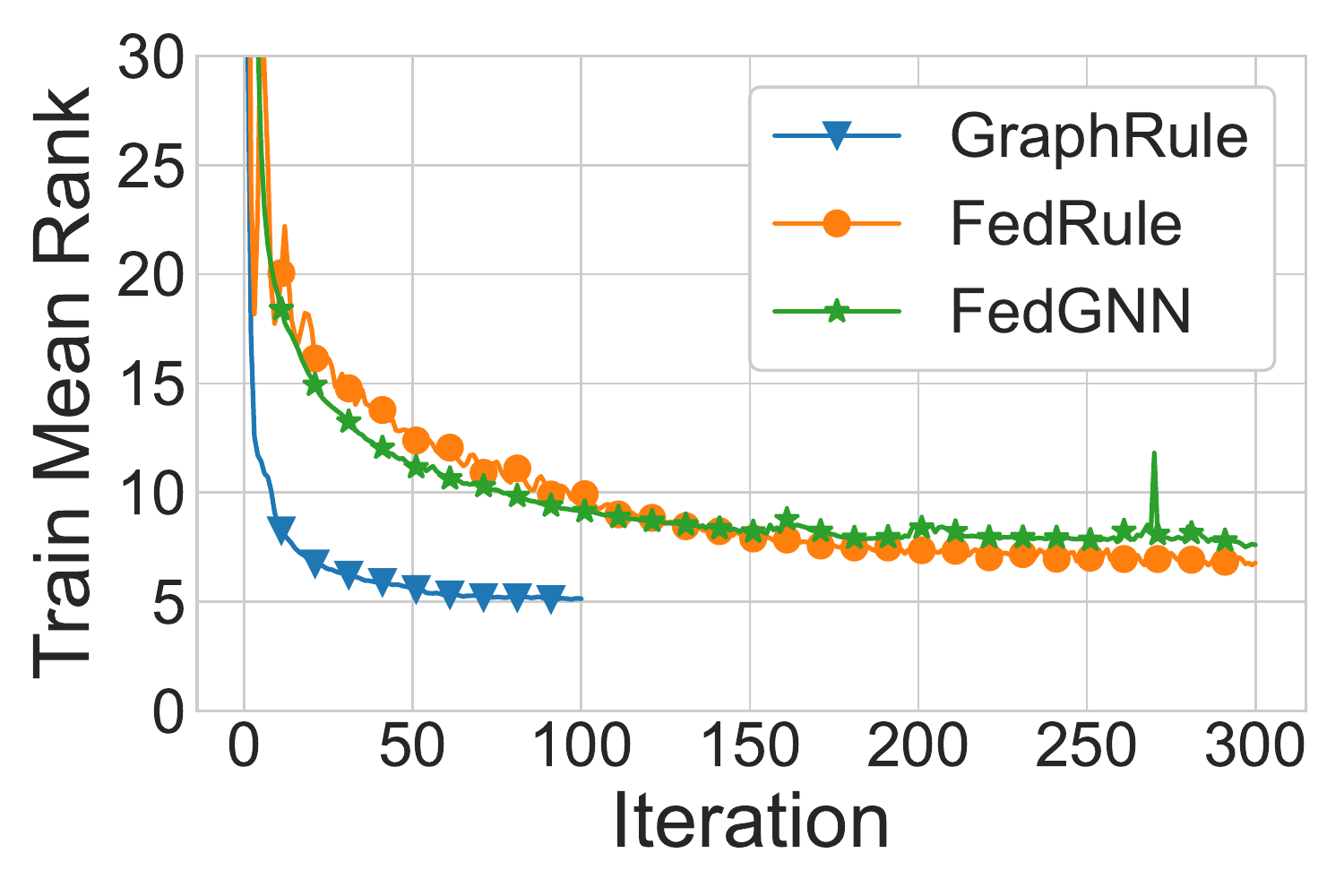}
    \includegraphics[width = 0.23\textwidth]{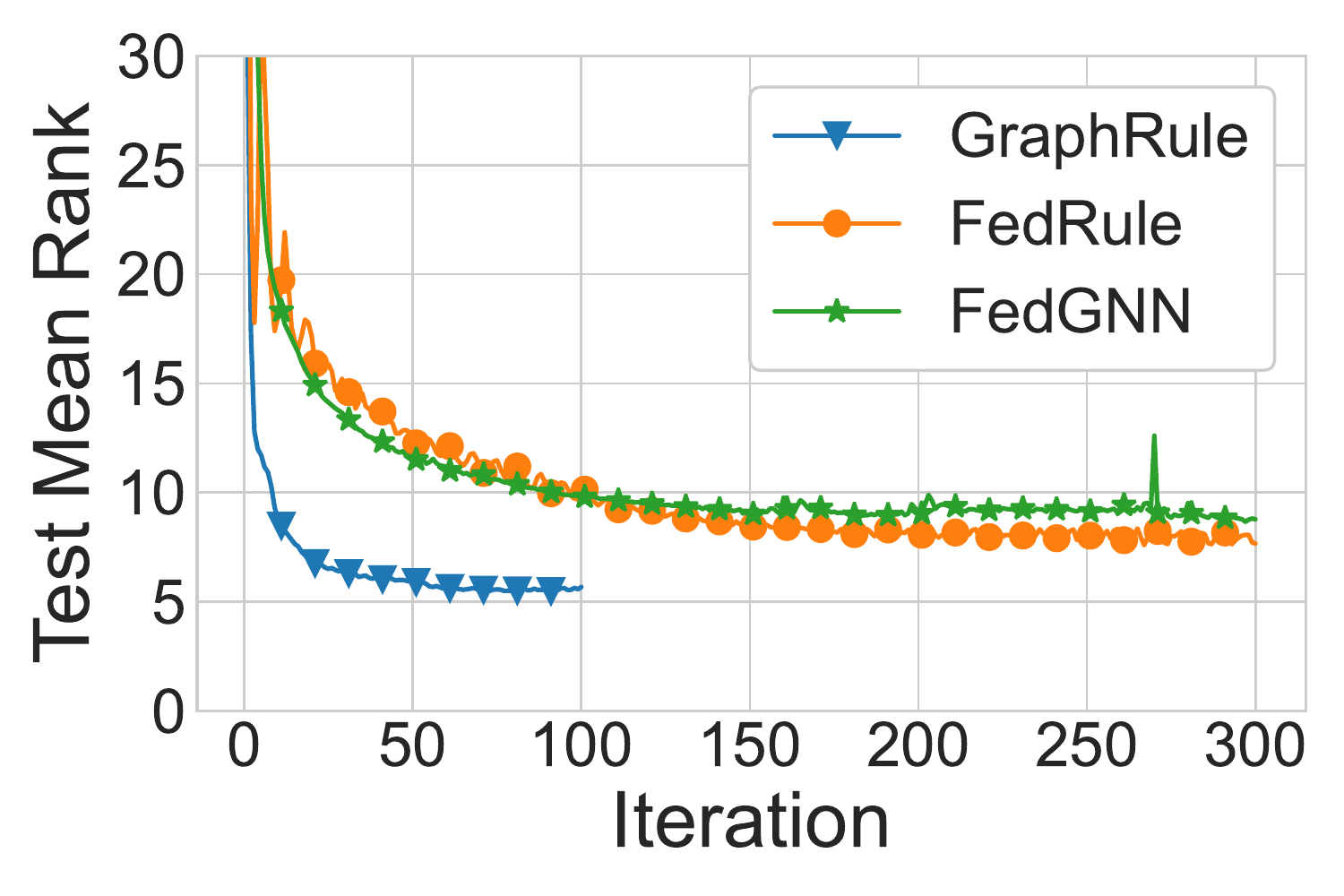}
    \caption{Training and test performance of different centralized and federated algorithms on Wyze Smart Home Rules Dataset, when considering multiple entities with the same type.}
    \label{fig:train_result_wyze_same_devices}
\end{figure}

\paragraph{Results on IFTTT Dataset}
The results of applying these methods on the IFTTT dataset are shown in Figure~\ref{fig:train_result_ifttt_no_same}. The IFTTT dataset is relatively smaller than the Wyze Smart Home Rules data, so it is easier to converge during the training process. For the training loss, the centralized training, \graphrule, convergences faster since it benefits from the IID distribution of data. For the federated setting, given it is a small dataset, the train loss of FedGNN converges faster than \fedrule with the same learning rate but the test loss of FedGNN diverges after $50$ iterations. Similar to the Wyze Smart Home Rules dataset, FedGNN has a better test AUC at the start but it decreases after $50$ iterations as the test loss diverges. The test AUC of \fedrule increases steadily during the training. Also, its test mean rank smoothly decreases, converging to an even lower value than the \graphrule. Table~\ref{fig:test_table_ifttt} shows the final test results on IFTTT data. Although \graphrule has a better testing loss and a slightly better AUC, the Mean Ranks of \fedrule are better than those of \graphrule, which is a better indicator of the recommendation performance.

Figure~\ref{fig:test_hit_rate_ifttt_no_same} shows the hit rate of different algorithms in the IFTTT dataset. Given the huge number of rules and apps, there are trigger-action pairs that are infeasible between some entities in practice. Since the number of users is small and hard to train a general model to avoid these infeasible pairs, we do rule filtering on the recommended rules to keep valid rules. Due to the small size of the dataset, the evaluation has high variance. FedGNN is slightly better than GraphRule and FedRule for the top-5 recommendation. But \fedrule is the best in general and has at the best $8.1\%$ higher hit rate than \graphrule and FedGNN.

\paragraph{Entities with Same Type}
\begin{figure}[t]
    \centering
    \includegraphics[width = 0.43\textwidth]{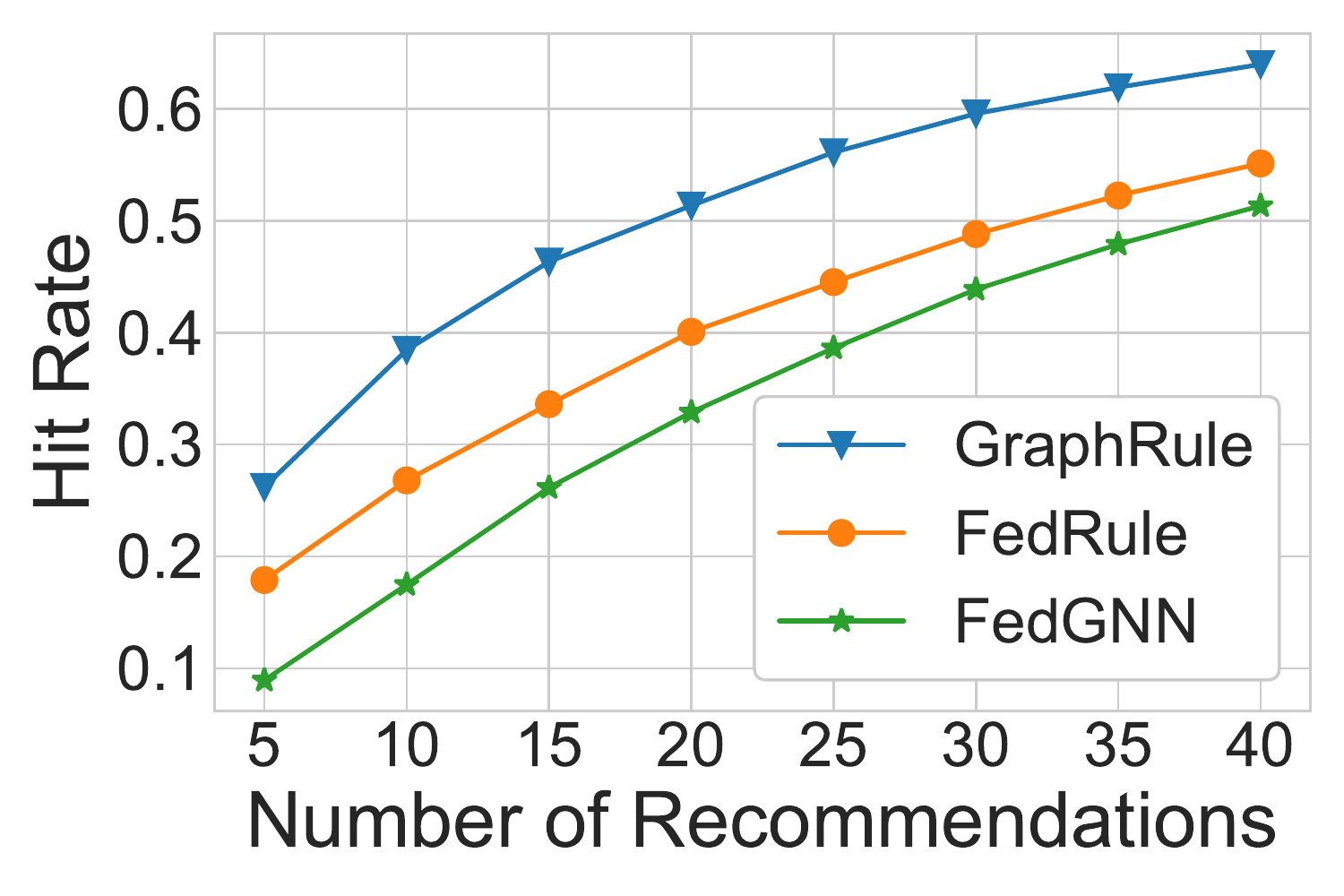}
    \caption{The Hit Rate on the test set for Wyze Smart Home Rules dataset, considering multiple entities with the same type. We compare \graphrule with \fedrule and FedGNN algorithms.}
    \label{fig:test_hit_rate_wyze_same_devices}
\end{figure}
As it was mentioned, the graph structure allows us to distinguish between different entities with the same type in a user's graph. 
In this case, user-item-based methods become infeasible. User graph-based methods, however, can solve this problem by simply considering these entities as different nodes in the graph and using a node embedding to distinguish between different nodes with the same type. Figure~\ref{fig:train_result_wyze_same_devices} shows the training and testing results for centralized and federated methods. Similar to the previous part, \graphrule converges faster, and FedGNN diverges after $150$ iterations. Similarly, the test loss of FedGNN diverges after $100$ iterations while \fedrule converges smoothly. Also, the AUC and Mean Rank results show that \fedrule has a better performance than FedGNN. Moreover, Figure~\ref{fig:test_hit_rate_wyze_same_devices} shows test hit rates. \graphrule has a better performance than federated approaches.

\subsection{GNN Model Selection}
\begin{figure}[bt]
    \centering
    \includegraphics[width = 0.23\textwidth]{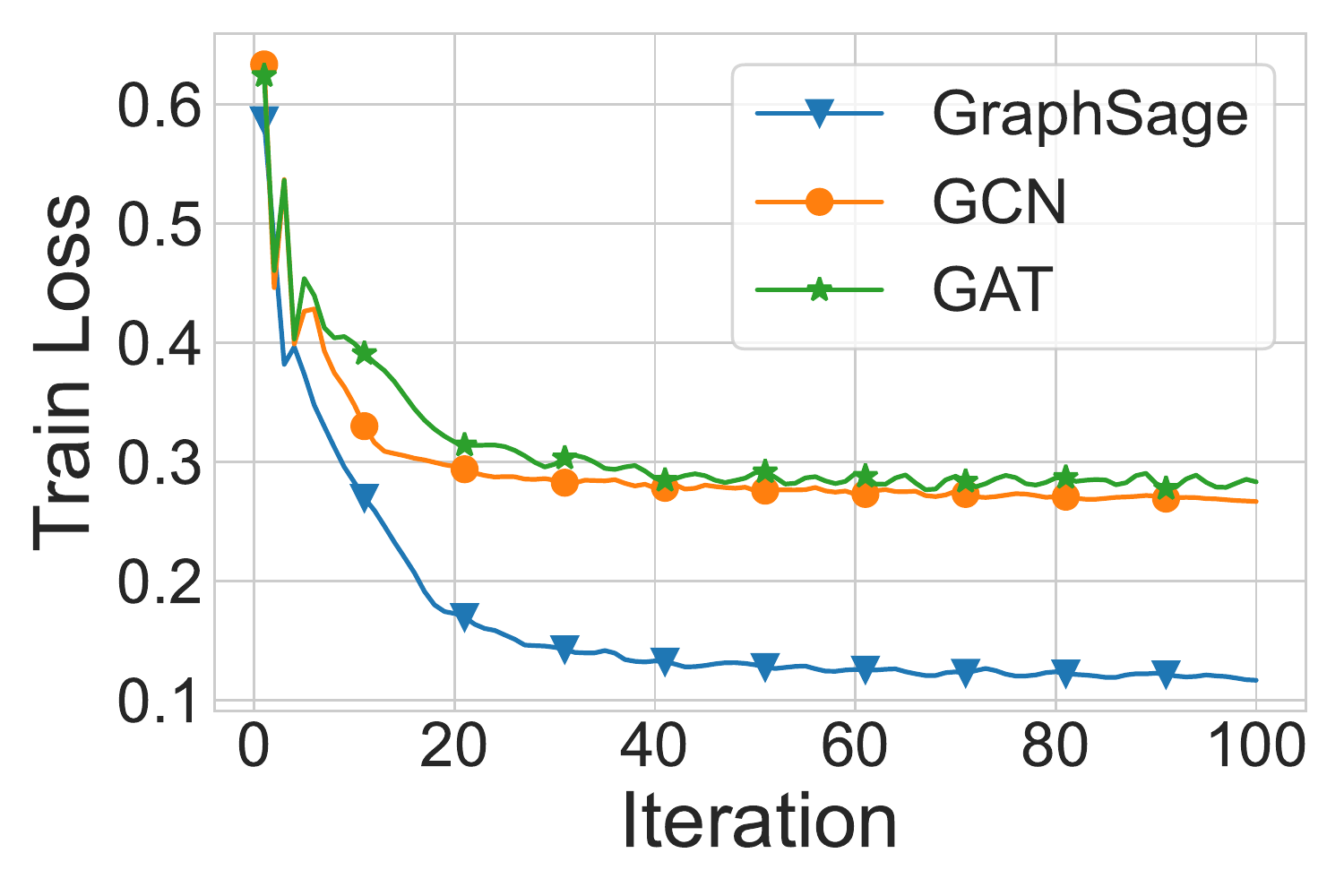}
    \includegraphics[width = 0.23\textwidth]{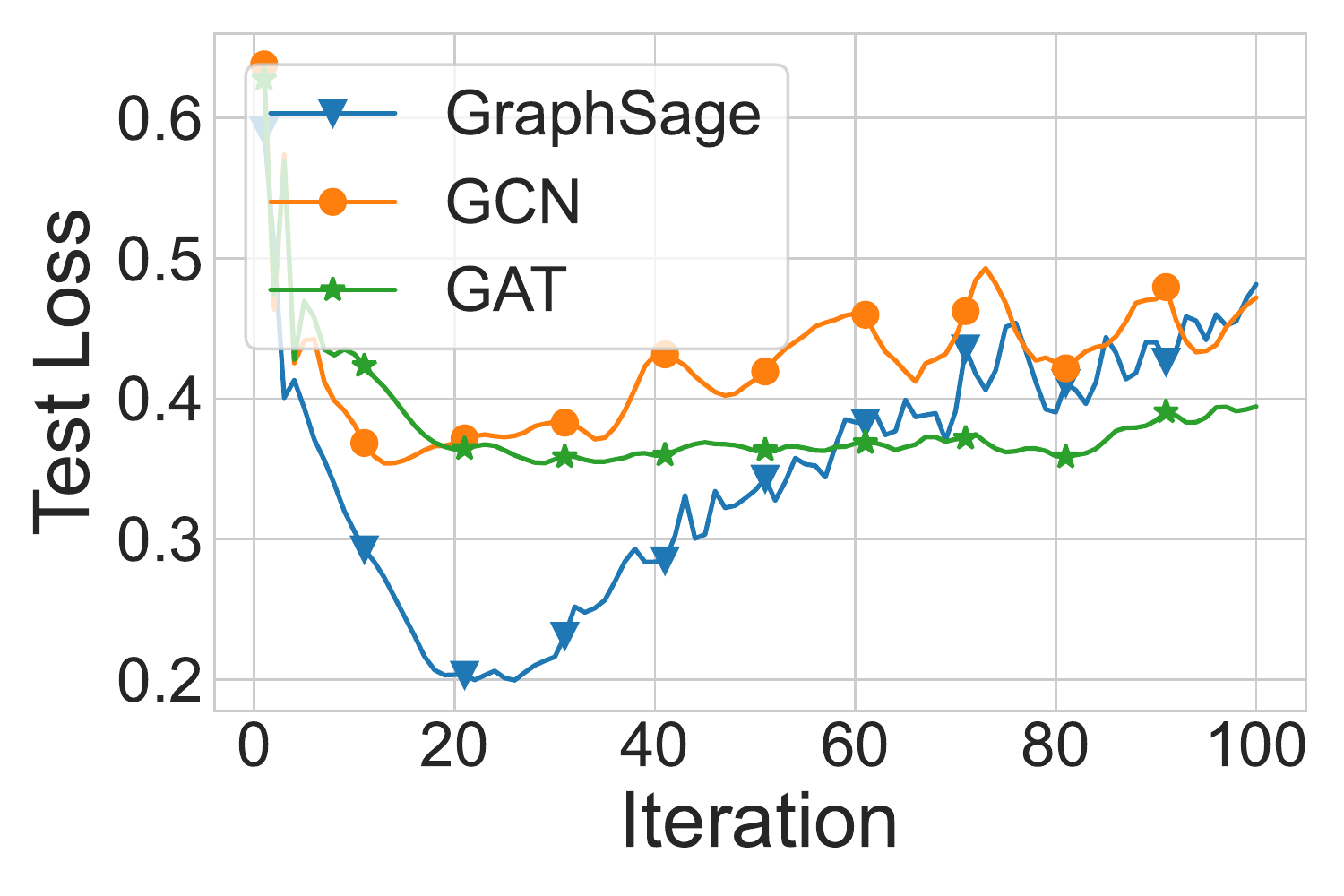}
    \includegraphics[width = 0.23\textwidth]{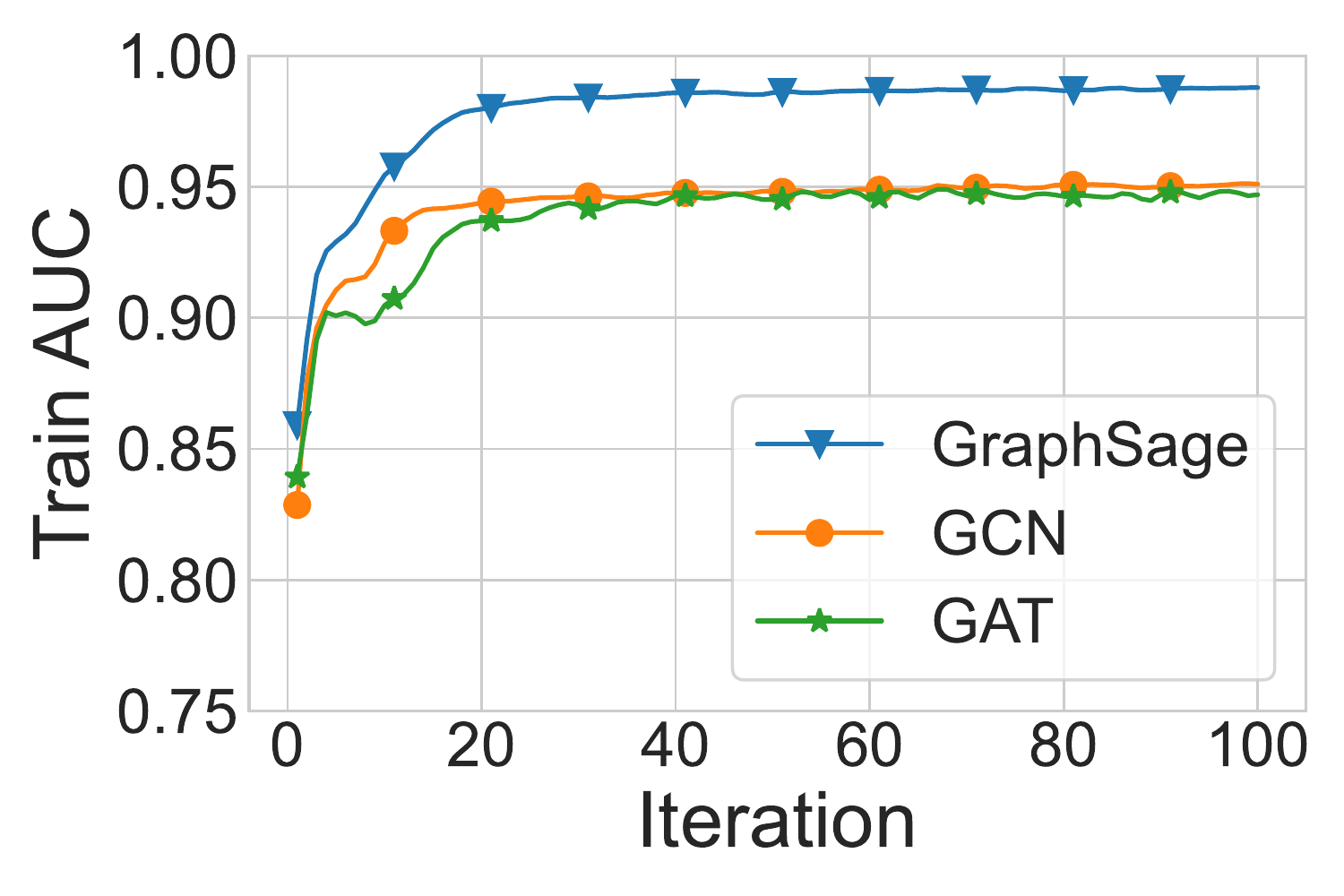}
    \includegraphics[width = 0.23\textwidth]{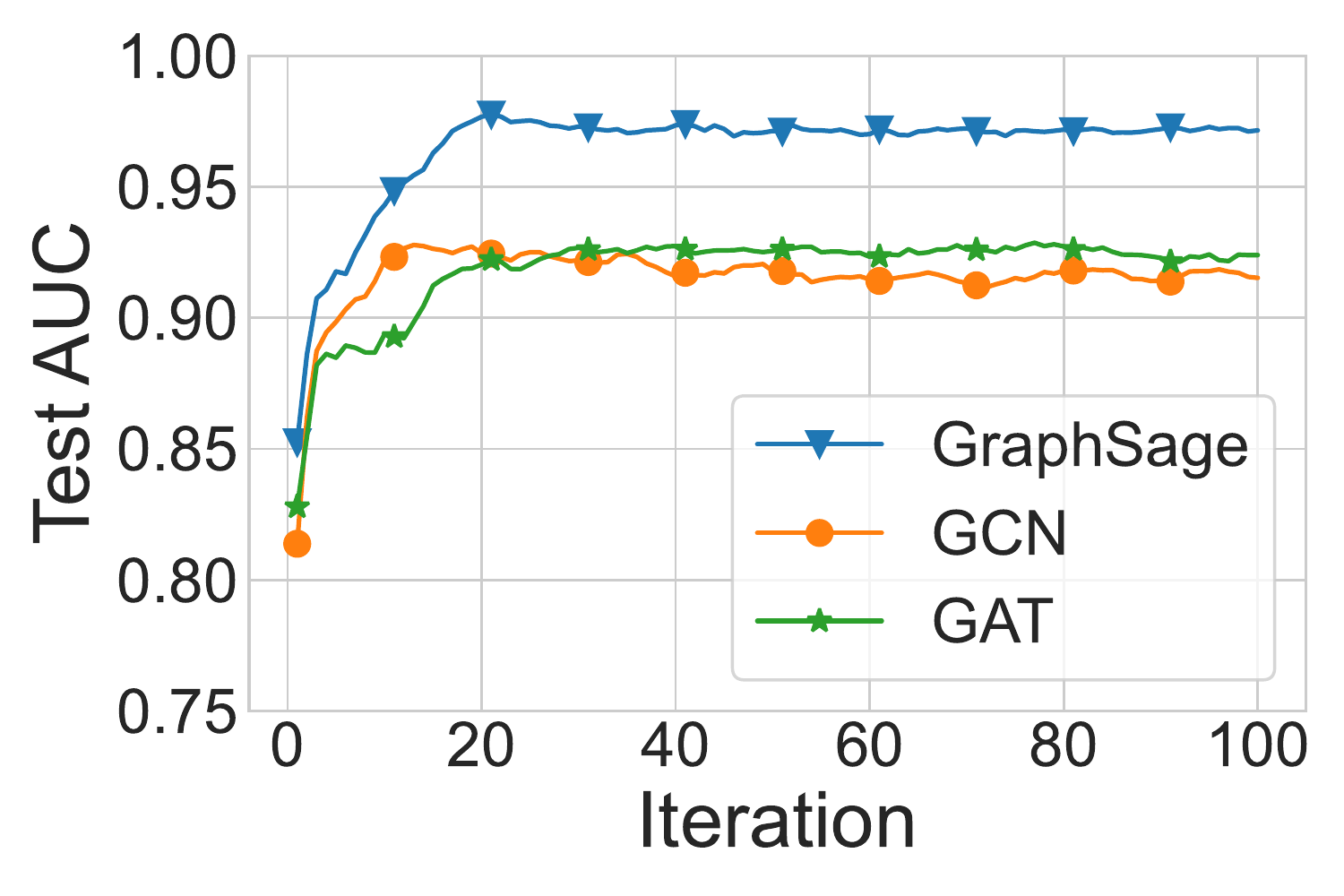}
    \includegraphics[width = 0.23\textwidth]{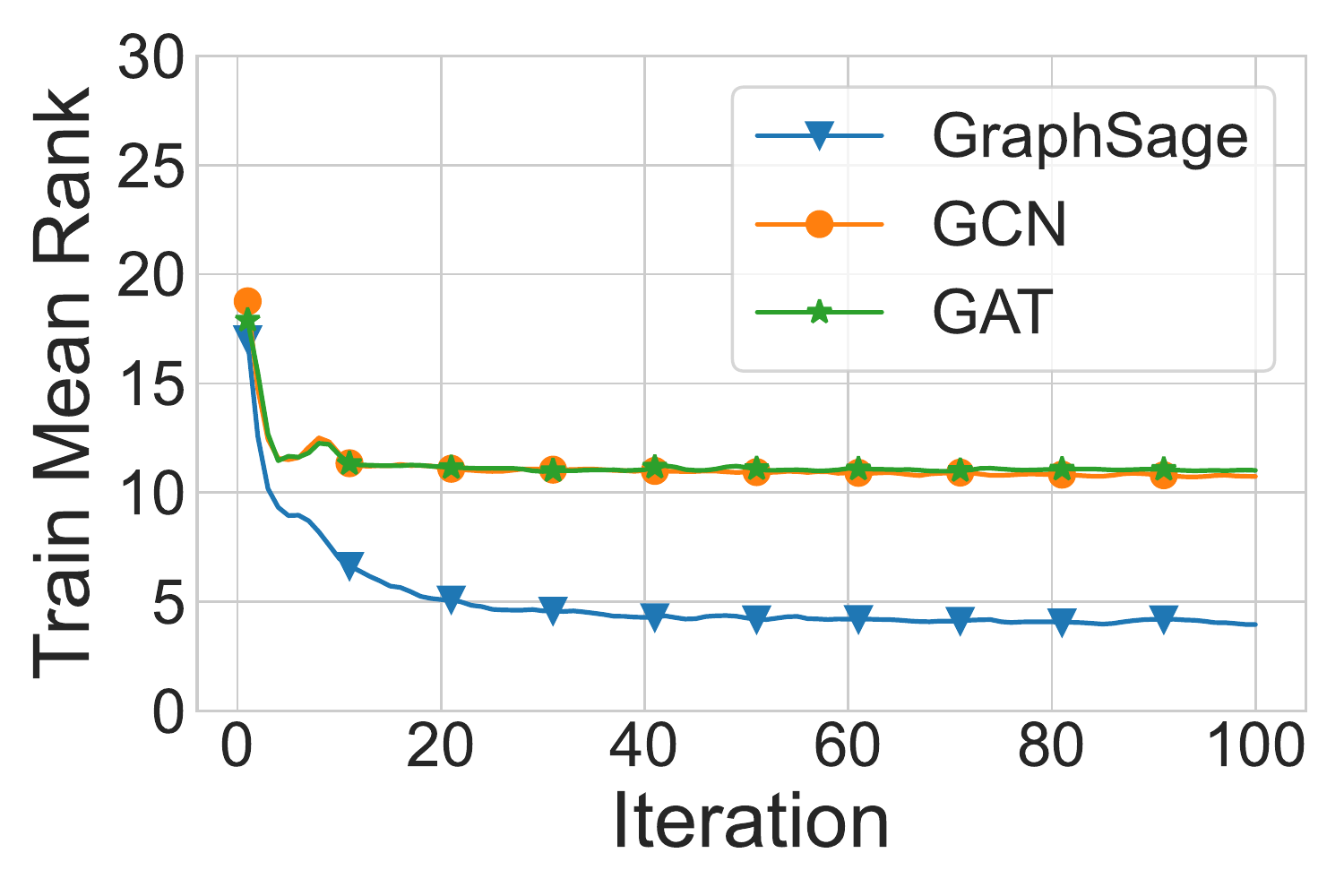}
    \includegraphics[width = 0.23\textwidth]{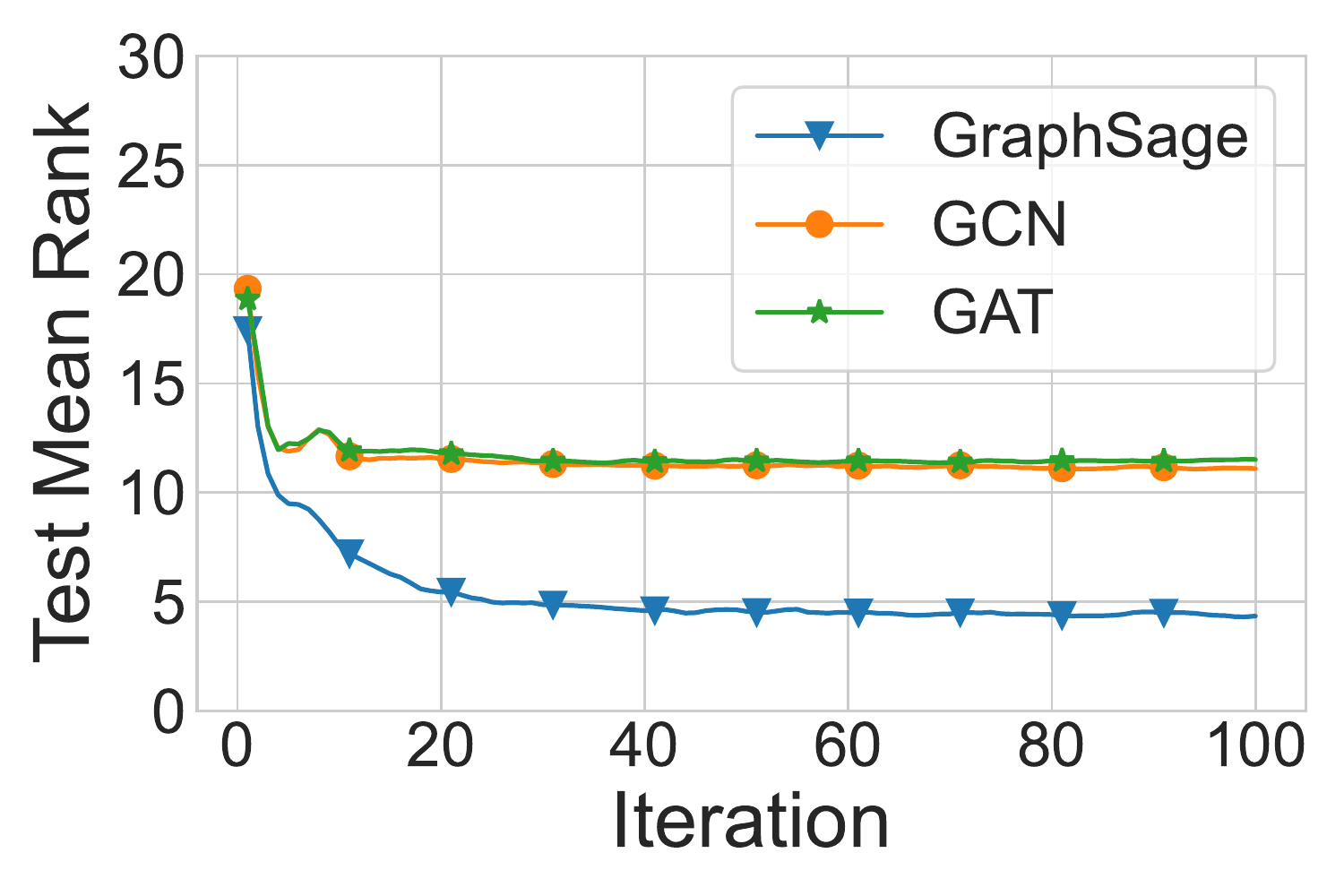}
    \caption{Train and test performance of different Graph Neural Network models on Wyze Smart Home Rules Dataset. GraphSage has better model performance than GCN and GAT.}
    \label{fig:train_result_wyze_no_same_gcn_graphsage_gat}
\end{figure}

To select the best model for our system we run an ablation study with two other GNN structures. We compare the GraphSage model with GCN~\cite{kipf2016semi} and GAT~\cite{velivckovic2017graph}. As shown in Figure \ref{fig:train_result_wyze_no_same_gcn_graphsage_gat}, GraphSage has better performance (loss, AUC, and mean rank) during training and testing. The main reason is the 0-in-degree nodes in the device graphs. Output for those nodes is invalid since no message will be passed to those nodes, which causes silent performance regression. As in Algorithm~\ref{Alg:graphsage}, for the 0-in-degree nodes, GraphSage concatenates its node feature and the neighbor features (the neighbor feature is a zero vector in this case) to avoid this issue.

\subsection{System Implementation}\label{sec:system}
We have implemented our rule recommendation in our smart home system, where users' devices and rules are represented as graphs. The system designed for this purpose has three major components, that are the data pipeline, the training API, and the inference API. For each user, the data pipeline connects to two databases, device ownership, and rules. By combining these two datasets, it can generate a complete graph for each user based on their available rules and devices (even those devices that have not been set up for any rules yet). The data pipeline functionalities are mainly the same for both the training and inference processes. It also generates a 3-dimensional tensor for valid rules based on source and target node types. This tensor will be used to heavily boost the inference time by about 20 times reduction when searching for valid rules. In the training process for the \graphrule, we generate the graphs for each user and keep 20\% of the rules as the test cases to avoid overfitting. The inference API is written to be called either by each user independently or with a bulk of users in parallel. When the app from the user side calls the inference API, first the data pipeline generates the user graphs, and then the latest trained models are called to run the inference. 

\begin{figure}[bt]
    \centering
    \subfigure[]{
    \includegraphics[width = 0.42\textwidth]{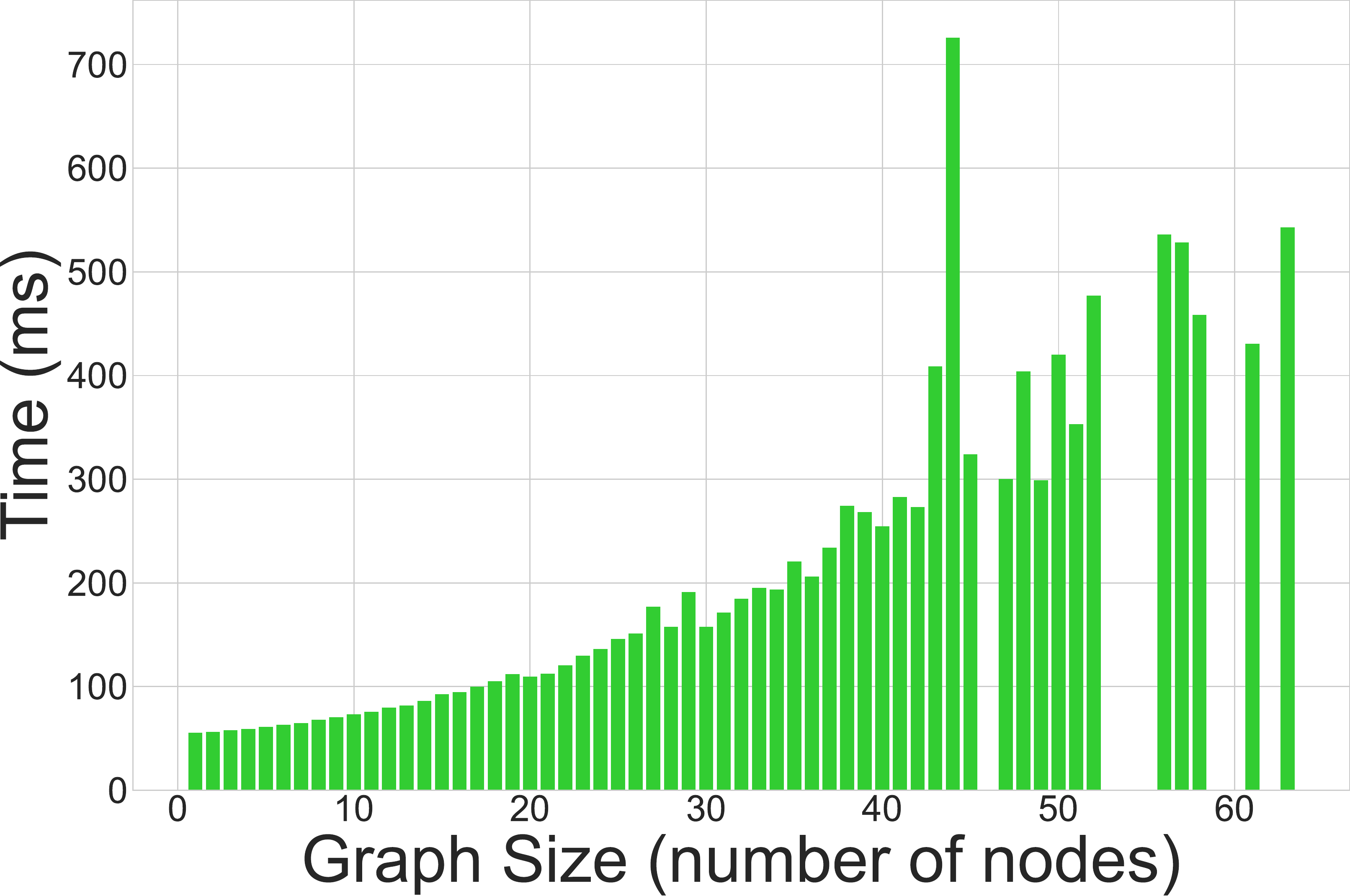}
    \label{fig:inference_time}}
    
    \subfigure[]{
    \includegraphics[width = 0.42\textwidth]{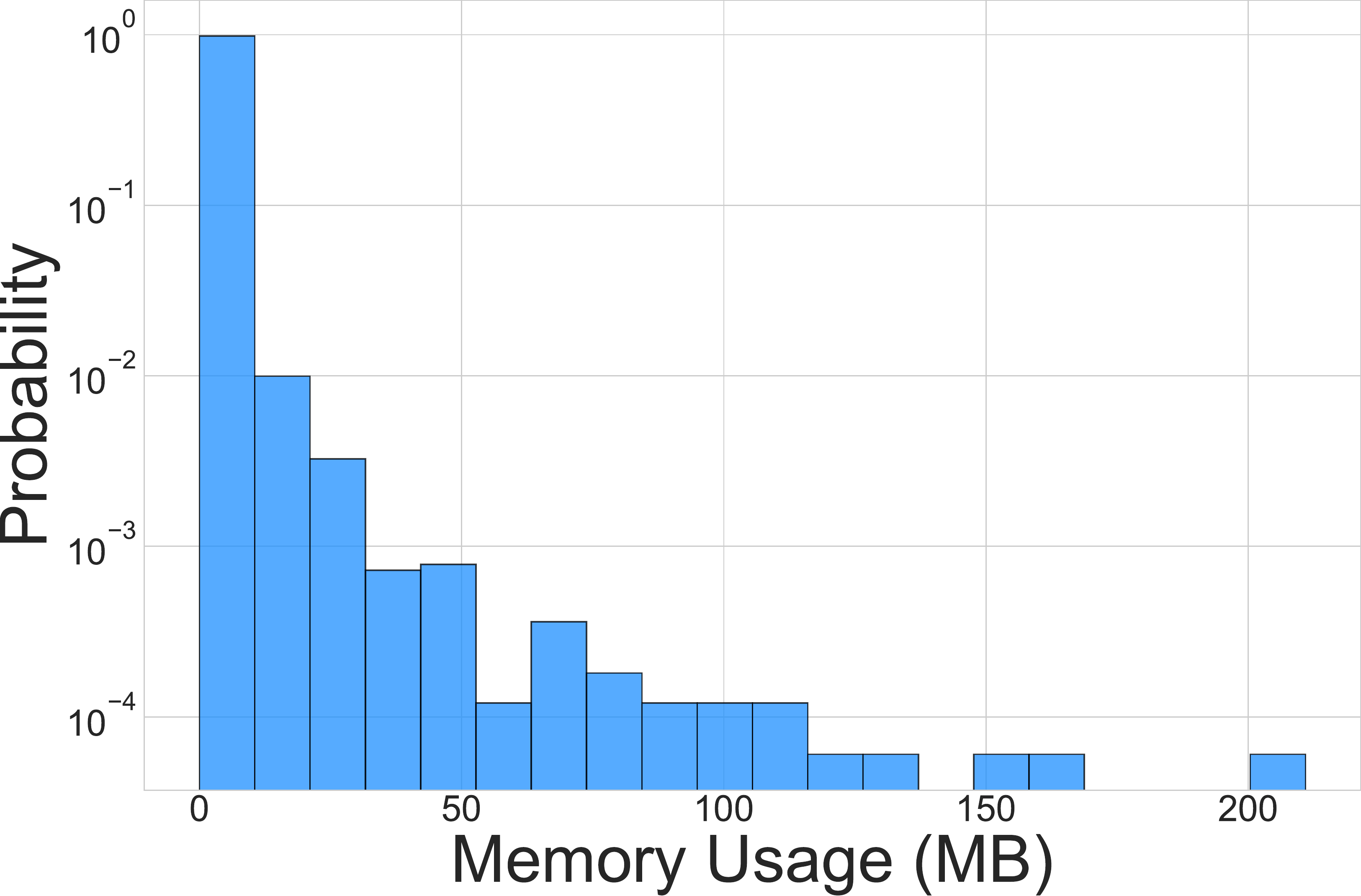}\label{fig:inference_memory}}
    \caption{The system performance of the Inference API for rule recommendation system. Figure (a) shows the delay time of the inference based on the size of the user's graph. Figure (b) shows the histogram of the memory used for the inference for various users. Most of the users got their recommendations with less than 10 MB of memory used and in less than 500 ms.}
    \label{fig:inference}
\end{figure}

Figure~\ref{fig:inference_time} and~\ref{fig:inference_memory} represent the performance of the API model on a \texttt{c5.4xlarge} instance of Amazon Web Services (AWS). This instance is configured on an Intel$^\text{\textregistered}$ Xeon$^\text{\textregistered}$ Platinum CPU with 3.00GHz frequency and 32~GB of RAM. In Figure~\ref{fig:inference_time}, the time required to complete the inference task from generating the graph to generating new recommendations based on the size of the user's graph is depicted. It is evident that by increasing the graph size the time required to complete the inference would increase. However, this increase seems to be linear, and for most users, it can be done in less than 500 ms on a normal machine. Figure~\ref{fig:inference_memory}, shows the memory usage histogram for different users. As it can be inferred almost all users' inference tasks can be done with as low as 10 MB in memory. This shows that this task can be easily done on edge devices on the user side.

\section{Conclusion}\label{sec:conclusion}
In this paper, we have presented a new formulation for the rule recommendation systems, called \graphrule, based on the graph structure of entities and their connections in a network of users. Despite the user-item format has been widely adopted in conventional recommendation systems, the graph structure is better at handling multiple entities with the same type and providing feasible recommendations based on the available entities in the user's graph. In addition, this graph structure enables us to have local models for each user and hence can further improve the privacy of users. We therefore further propose \fedrule, a federated rule recommendation setting based on the graph structure. When compared with the previous proposals for federated learning on graphs, \fedrule is designed for cross-device federated learning to ensure each user's data is kept private locally. Due to the variance reduction mechanisms in \fedrule, we are able to overcome the Non-IID issue of prior approaches in highly heterogeneous networks of users.

Extensive experimental results on Smart Home Rules and IFTTT datasets present the effectiveness of the proposed formulation and training approaches over prior proposals. Providing more personalized recommendations based on each user's behavior in the federated learning setup is left as a future direction. Another direction of future work might be to learn further about privacy: federated learning models can sometimes be reverse-engineered to reveal private information about a user. To provide a more insightful comparison, real-world implementation of the proposed system and investigating it with rigorous A/B testing is necessary and will be also considered as future work.

\section*{Acknowledgement}
The authors would like to acknowledge that the icons used in this paper are made by ``Flaticon'', ``Justicon'', ``Nikita Golubev'', and ``Good Ware'' from \url{www.flaticon.com}.
\bibliography{example_paper}
\bibliographystyle{ACM-Reference-Format}



\end{document}